\documentclass[11pt]{article}

\usepackage[final]{acl}
\usepackage{booktabs} 
\usepackage{times}
\usepackage{latexsym}

\usepackage[T1]{fontenc}

\usepackage[utf8]{inputenc}

\usepackage{microtype}

\usepackage{inconsolata}

\usepackage{graphicx}

\usepackage{subcaption} 
\usepackage{float} 
\usepackage{multirow} 
\usepackage{array}     
\usepackage{tabularx}  
\usepackage{caption}
\usepackage{array}
\usepackage{booktabs}
\usepackage{enumitem}
\usepackage{mathrsfs}
\title{Why Knowing Both Hops Is Not Enough:
Understanding Two-Hop Generalization in Language Models}

\author{
  \textbf{Zili Zhang}\textsuperscript{1}\thanks{Equal contribution.} \quad
  \textbf{Yilin Wang}\textsuperscript{1}\footnotemark[1] \quad
  \textbf{Heng Wang}\textsuperscript{2} \quad
  \textbf{Herun Wan}\textsuperscript{1} \quad
  \textbf{Minnan Luo}\textsuperscript{1,3}\thanks{Corresponding author.} \\
  \textsuperscript{1}Xi'an Jiaotong University \quad
  \textsuperscript{2}University of Illinois Urbana-Champaign \\
  \textsuperscript{3}National Engineering Research Center for Visual Information and Applications, China \\
  \texttt{\{zhangzl666, 13148035071xjtu\}@stu.xjtu.edu.cn}
}

\usepackage{amsmath}
\begin{document}
\maketitle
\begin{abstract}
Large language models (LLMs) can solve complex multi-hop problems yet exhibit puzzling failures on simple two-hop queries: although a model may correctly store each individual hop, it often fails to combine them. 
To understand the internal mechanisms of this phenomenon, we train transformers from scratch in a controlled symbolic environment.
Our experiments reveal a pattern in two-hop generalization: models generalize reliably when the second hop follows the training distribution, but always fail when it deviates. Through mechanistic analysis, we provide a complete explanation for these distinct generalization behaviors: in settings where models generalize successfully, performance is driven by the emergence of consistent intermediate representations for the same entities across contexts, whereas failures on settings where the second hop is out-of-distribution arise from a mismatch across layers: lower layers correctly construct these intermediate representations, but upper layers, while trained on corresponding atomic facts, primarily learn to map them to outputs rather than to reason over them. Driven by this insight, we propose a recurrent-style training strategy, which enables transformers to reuse their reasoning circuitry across input forms and substantially improves generalization on out-of-distribution two-hop queries. Our data and code are available at \url{https://github.com/zzl-strong/two_hop}.

\end{abstract}

\section{Introduction}
Large language models (LLMs) have achieved impressive performance across multiple tasks \citep{zhaoSurveyLargeLanguage2025}, yet their ability to perform multi-hop reasoning remains fragile and poorly understood \citep{huang-chang-2023-towards}. Even in the simplest case—two-hop reasoning—models exhibit a persistent failure: although they can correctly store each atomic fact (e.g., A $\to$ B and B $\to$ C), they often struggle to reliably combine them into a compositional inference (A $\to$ C) \citep{biran-etal-2024-hopping,balesni2024lessons}.

Therefore, enhancing the ability of multi-hop reasoning has recently become a research focus \citep{li-etal-2024-understanding, openai2024o1, petty-etal-2024-impact, Luo2024ImproveMR}. Beyond explicit prompting strategies such as Chain-of-Thought (CoT) \citep{kojima2022large, wei2022chain}, a growing line of work suggests that LLMs can answer compositional queries without generating intermediate steps, indicating the presence of implicit multi-hop reasoning \citep{deng2024explicit, yang2024large}. However, the underlying mechanism governing how transformers internally complete two-hop queries using latent reasoning remains not well understood.


In this study, we aim to dissect the internal mechanisms of this phenomenon without the confounding factors of pretraining data opacity—such as simple memorization or spurious correlations \citep{xu-etal-2022-model, wang-etal-2023-causal, ju-etal-2024-investigating}. To this end, in \S\ref{section:2}, we use a symbolic environment to train a transformer from scratch and test our trained model on different splits of data.

Using this approach, in \S\ref{section:3}, we observe an asymmetric generalization pattern: models generalize reliably when the second hop follows the same distributional patterns observed during training (Test-II, Test-OI), but fail when the second hop is out-of-distribution (Test-IO, Test-OO). 

To give a complete explanation, we conduct a mechanistic analysis in \S\ref{section:4}. First, we observe a progressive reasoning process across layers using logit lens. Based on this experiment, we select layer 5 as the anchor for further analysis and define its output at position ${r_1}$ as the representation of the bridge entity. Then, through causal patching, we find that the alignment of the same entity's representations is positively correlated with the model's generalization on Test-II/OI. To explain the failure on Test-OO and Test-IO, we hypothesize that the upper layers of the model only learn the mapping ability during atomic facts training, and validate it by linear transformation and attention knockout experiments.

Driven by the above insights, in \S\ref{section:5}, we heuristically propose that the looped architecture is a way to solve multi-hop reasoning failure: the model eliminates the difference between training and inference by aligning representations across different layers. Extending experiments verify the scalability of our method and its effectiveness on real-world corpora.

In summary, this paper makes the following contributions:
\begin{itemize}
    \item \textbf{Identify a systematic asymmetry in two-hop generalization}: models generalize reliably when the second hop stays within the training distribution, and, conversely, fail when the second hop is out-of-distribution, despite correct atomic knowledge.
    \item \textbf{Provide a complete mechanistic explanation for the two-hop phenomenon}: we attribute the generalization success on Test-II/OI to the same entity's representation alignment, and its failure on Test-IO/OO to a functional mismatch where upper layers specialize in representation mapping rather than reasoning.
    \item \textbf{Propose a mechanism-grounded solution}: we adopt a looped architecture and show that aligning representation formats across layers enables robust two-hop generalization. Extending experiments show that our method is also applicable to modern architectures and real-world language data.
\end{itemize}

\section{Experimental Settings}
\label{section:2}
We construct the data in a symbolic environment for controlled analysis, following \citet{wang2024grokking}.

\begin{figure}[ht]
    \vspace{-2mm}
    \centering
    \includegraphics[width=\linewidth]{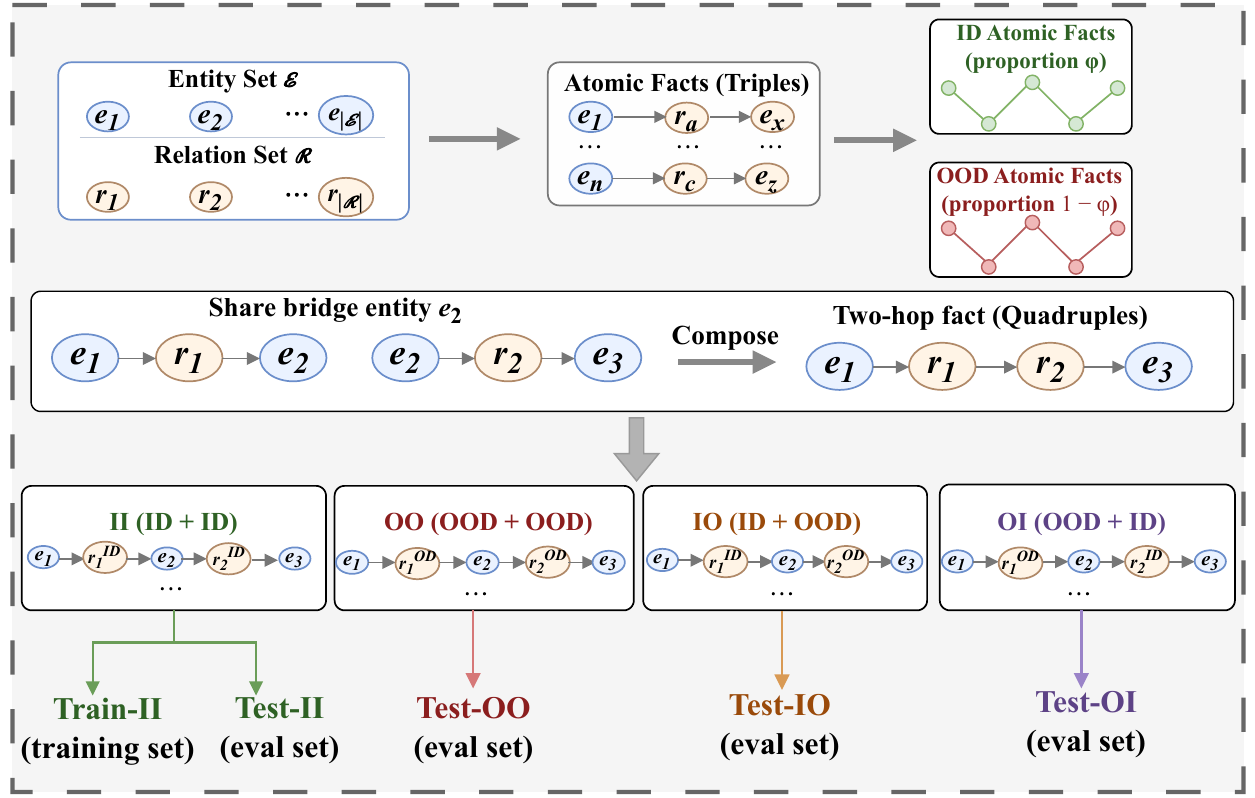}
    \caption{The symbolic data construction pipeline.}
    \vspace{-3mm} 
    \label{fig:data_const}
\end{figure}

\textbf{Data Construction:} The overall construction pipeline is shown in Figure \ref{fig:data_const}: We define an entity set $\mathcal{E}=\{e_1,\ldots,e_{|\mathcal{E}|}\}$ and a relation set $\mathcal{R}=\{r_1,\ldots,r_{|\mathcal{R}|}\}$. For each head entity $e_1\in\mathcal{E}$, we sample 20 relations from $\mathcal{R}$ and pair each with a randomly chosen tail entity $e_2\in\mathcal{E}$, yielding atomic facts $(e_1,r,e_2)$. We partition atomic facts into in-distribution (ID) and out-of-distribution (OOD) subsets with proportion $\phi$. We then form two-hop facts $(e_1,r_1,r_2,e_3)$ by chaining $(e_1,r_1,e_2)$ and $(e_2,r_2,e_3)$ that share bridge entity $e_2$. These facts fall into four types: II, OO, IO, and OI. II is further split into Train-II and Test-II, while OO/IO/OI are used as Test-OO/Test-IO/Test-OI. The cross-distribution splits follow \citet{ye2026transformers}. Details are in Appendix \ref{appendix:dataset-details}.

\textbf{Training Configuration:} Our training split consists of all atomic facts (ID and OOD) and Train-II. All other two-hop splits (Test-II, Test-IO, Test-OI, and Test-OO) are used for evaluation to comprehensively assess the generalization ability of the model. We train a GPT2-style decoder-only transformer with 8 layers and a hidden dimension of 768. The model is optimized using AdamW with a learning rate of $1\times10^{-4}$. More detailed training configuration is provided in appendix \ref{appendix:exp-env}.

\section{Training Results}
\label{section:3}
\begin{figure*}[t]
\centering
\includegraphics[width=\textwidth]{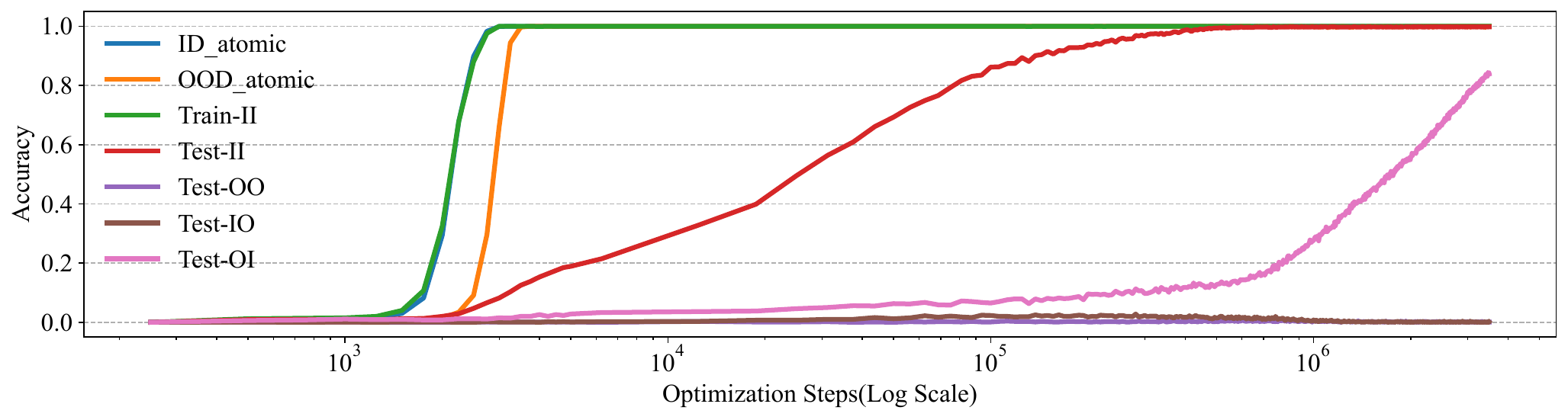}

\caption{Training dynamics of a transformer trained on all atomic facts and in-distribution two-hop compositions. Accuracy is shown for atomic facts, the two-hop training set (Train-II), and multiple evaluation splits. The model rapidly fits the training data, followed by delayed generalization to the in-distribution test set (Test-II) and partial improvement on Test-OI. In contrast, performance on Test-IO and Test-OO remains near chance throughout training, indicating a failure to generalize when the second hop is out-of-distribution.}
\label{fig:training_results}

\end{figure*}

Figure \ref{fig:training_results} shows the training dynamics across different data splits. We observe that: \textbf{(A)} The model first achieves near-perfect accuracy on Train-II after a relatively small number of optimization steps. \textbf{(B)} After continued training, the accuracy on Test-II begins to increase steadily and eventually reaches perfect accuracy, suggesting that the model transitions from memorization to generalization. \textbf{(C)} At a later stage of training, the model also exhibits noticeable gains on the Test-OI split, indicating the emergence of an additional generalization phase. \textbf{(D)} However, even with longer training, the model cannot generalize on Test-IO and Test-OO splits.

Overall, the model exhibits a highly asymmetric generalization pattern: it generalizes reliably only when the second hop stays within the training distribution (Test-II and Test-OI), and consistently fails when the second hop is out-of-distribution (Test-IO and Test-OO). In the following section, we investigate the underlying reasons for this disparity by analyzing how intermediate representations are formed and processed during two-hop reasoning.

\section{Mechanism of Two-Hop Reasoning}
\label{section:4}
\subsection{Identifying the Bridge Entity Representation in Hidden Space}
\label{section:4.1}

To explain the asymmetric training dynamics observed above, we begin by identifying where and how the bridge entity $e_2$ is represented in the model. 
To this end, we conduct a logit lens \citep{nostalgebraist2020logitlens} analysis on the trained model at both the $r_1$ and $r_2$ positions. 
At the $r_1$ position, we track how the probability of $e_2$ evolves across layers, probing when the model acquires sufficient information about the intermediate entity. 
At the $r_2$ position, we examine the probabilities of the relation token $r_2$ itself and the target entity $e_3$, which together reflect the beginning of second-hop reasoning.

\begin{figure}[ht]
    \centering
    \includegraphics[width=\linewidth]{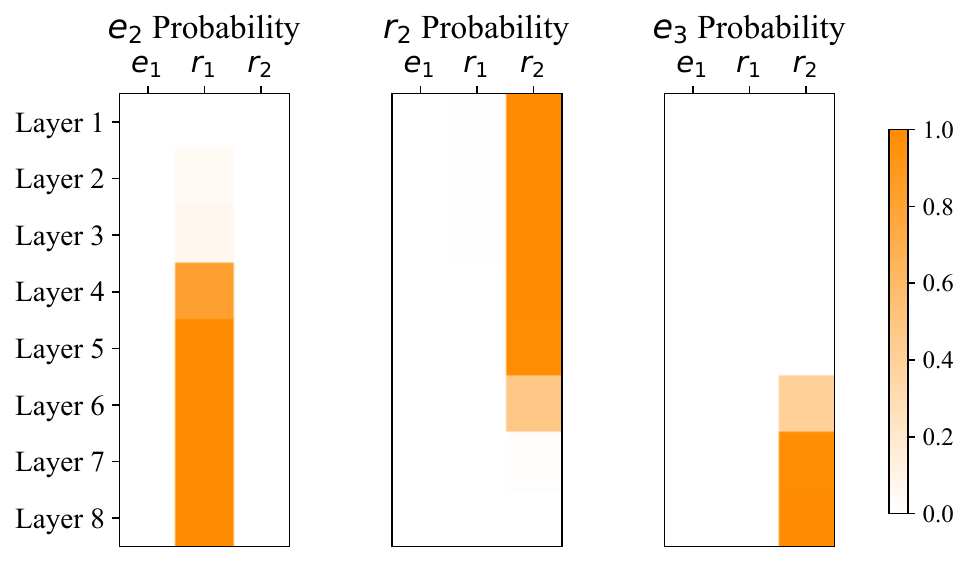}
    \caption{Logit lens reveals a progressive reasoning process, and layer 5 is a transition layer between two hops.}
    \vspace{-3mm} 
    \label{fig:logit_lens}
\end{figure}

As shown in Figure \ref{fig:logit_lens}, layer 5 emerges as a critical transition layer. 
By this layer, the model has already accumulated strong evidence for the bridge entity $e_2$, as indicated by the high probability of $e_2$ at the $r_1$ position. 
More importantly, after layer 5, we observe a qualitative shift in behavior at the $r_2$ position: the probability of the surface token $r_2$ begins to decrease, while the probability of the final answer $e_3$ rises, suggesting upper layers may start integrating contextual information and performing the second hop of reasoning.

This progressive reasoning process leads us to conjecture that the output at position $r_1$ of layer 5 may contain information about the bridge entity $e_2$ that is necessary for the model's reasoning. Therefore, we define it as the hidden representation of the bridge entity, denoted by $\mathscr{R}$, which serves as an anchor for subsequent analysis\footnote{In Appendix \ref{appendix:logit_lens_details}, we repeat this experiment at other training stages and find that the key concept layer only changes in the early memorization stage and is stable thereafter, so using this layer as the analysis anchor is reasonable.}.


\subsection{Probing Bridge Entity Representation Consistency via Entity Patching}
\label{section:4.2}

\begin{figure*}[t]
\centering
\includegraphics[width=\textwidth]{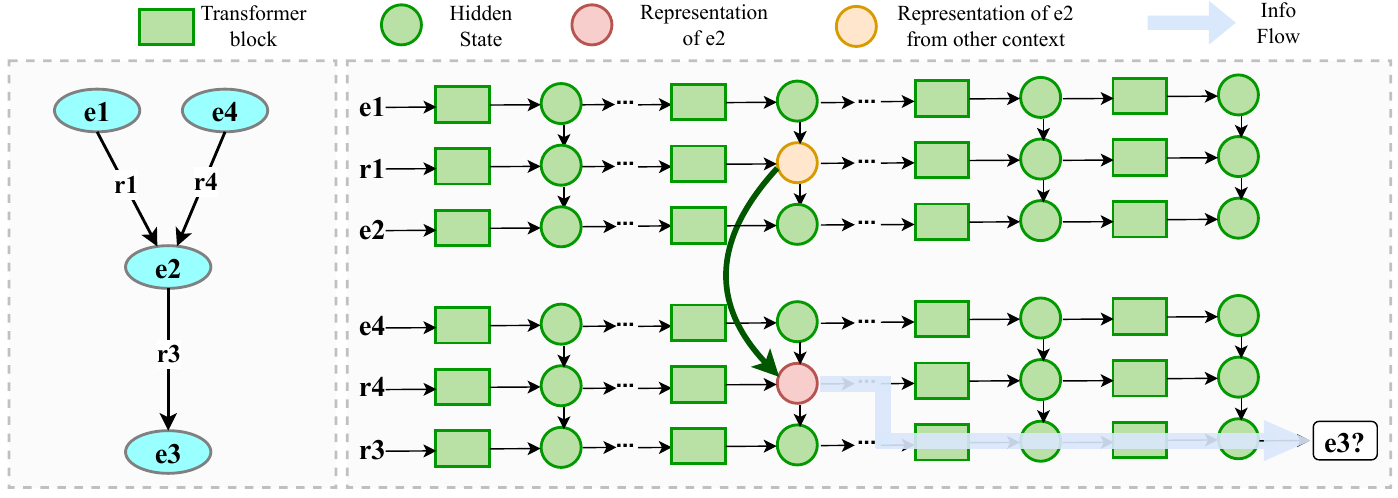}
\caption{\textbf{(Left)} A two-hop query composed of two single-hop facts, where the bridge entity $e_2$ connects $(e_4, r_4)$ and $(e_2, r_3)$ to yield the target entity $e_3$. \textbf{(Right)} The entity patching procedure: the hidden representation of the bridge entity $e_2$ at layer 5 is replaced with an alternative representation of the same entity extracted from a different single-hop context (e.g., $(e_1, r_1)$). The model’s ability to recover the original answer after patching tests whether representations of the same entity are aligned across contexts.}
\label{fig:patch}

\end{figure*}

To perform robust multi-hop reasoning, models must move beyond surface-level pattern matching and internalize the underlying graph structure of knowledge. A fundamental requirement for such capability is concept invariance: the internal representation of a bridge entity $e_2$ should remain consistent regardless of the specific single-hop context that elicits it. Based on this intuition, we hypothesize that the model's ability to generalize on Test-II and Test-OI is driven by the emergence of such consistent representations at the critical bridge layer. 

To test this hypothesis, we introduce an \textbf{entity representation consistency patching experiment}, illustrated in Figure \ref{fig:patch}. We consider a two-hop query from the in-distribution training set (Train-II), where the bridge entity $e_2$ appears at position $r_1$ and is represented by the hidden state at layer 5, following the definition established in Section \ref{section:4.1}. In this instance, the single-hop context that gives rise to $e_2$ is $(e_4, r_4)$. 

We then select another single-hop atomic fact with the same tail entity $e_2$ but a different context, e.g., $(e_1, r_1)$. The representation $\mathscr{R}_{e_2}$ extracted from this context is used to replace the original bridge entity representation in the two-hop query, while keeping all other components unchanged\footnote{To rule out $e_3$ being pre-encoded at $r_3$ in early training, we replace $r_3$ with that from another two-hop query whose answer is not $e_3$, so predicting $e_3$ must use the patched representation rather than residual earlier-layer information.}. After patching, we evaluate whether the model still produces the correct final answer $e_3$.

\begin{figure}[t]
    \centering
    \includegraphics[width=\linewidth]{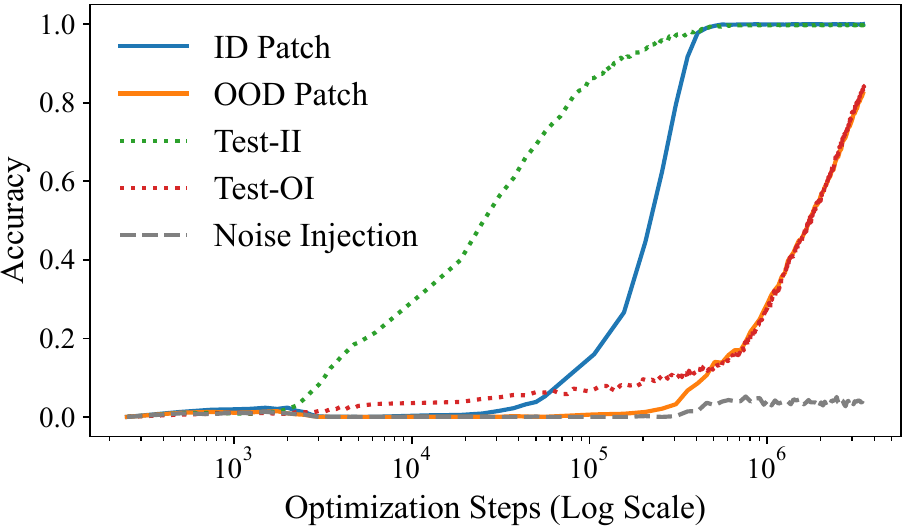}
    \caption{Patching success rate using ID and OOD bridge entity representations during training, compared with Test-II and Test-OI accuracy.}
    
    \label{fig:patch_result}
\end{figure}

We perform this replacement using bridge entity representations drawn from two sources: (i) in-distribution (ID) atomic facts and (ii) out-of-distribution (OOD) atomic facts, and record the patching success rate throughout training. Intuitively, if the model has learned a context-invariant representation for $e_2$, substituting its representation from another context should preserve the correctness of the second-hop inference. As a control, we also try to inject Gaussian noise, destroying the original representation to verify that this representation is causally necessary for the final prediction; the resulting performance drop (The grey line in Figure \ref{fig:patch_result}) ensures that patching success stems from representation alignment rather than the model ignoring the position.

We show the success rate of the patching experiment and the model's performance on Test-II and Test-OI in Figure \ref{fig:patch_result} as they change during the training process, which reveals a clear training-dependent pattern. In the early stages of training, patching success rates are low for both ID and OOD replacements, indicating that representations of the same entity remain highly context-dependent. During this phase, the model operates in a rote memorization regime, encoding entity information in a fragmented and inefficient manner.

As training progresses, the success rate of ID-based patching increases first. Notably, this rise coincides with the improvement of performance on Test-II, suggesting that generalization within the in-distribution setting emerges once representations of the same entity become aligned across different ID contexts. With further training, the success rate of OOD-based patching also begins to increase, accompanied by a corresponding improvement on Test-OI.

These results support the assumption that generalization in two-hop reasoning is driven by the gradual alignment of entity representations across contexts. Moreover, representations derived from in-distribution atomic facts converge more rapidly than those from out-of-distribution facts, explaining why generalization to Test-II precedes that to Test-OI.
\begin{figure}[ht]
    \centering
    \includegraphics[width=\linewidth]{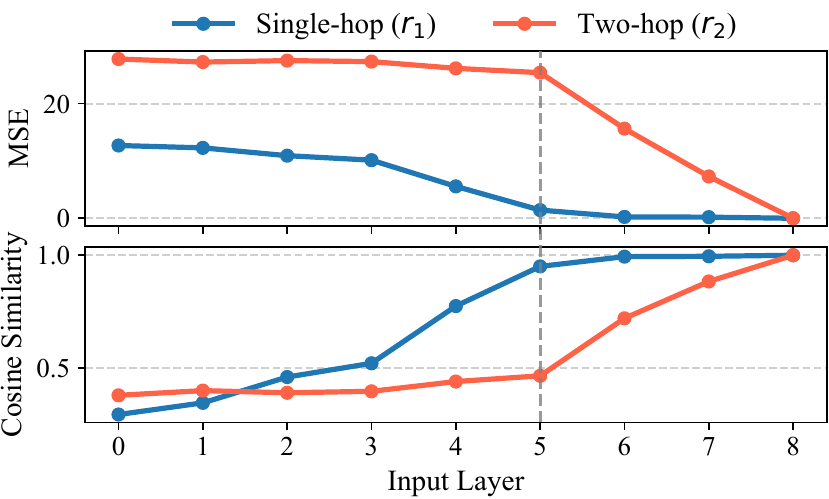}
    \caption{High cosine similarity and low MSE from layer 5 onward in single-hop queries indicate that upper layers act as linear mappers, whereas the lower linearity in two-hop queries suggests nonlinear information aggregation at the bridge layer.}
    \vspace{-3mm}
    \label{fig:linear_prediction}
\end{figure}
\subsection{Why Second-Hop OOD Fails: Reasoning vs. Mapping in Upper Layers}
\label{section:4.3}
\begin{figure*}[t]
\centering
\includegraphics[width=\textwidth]{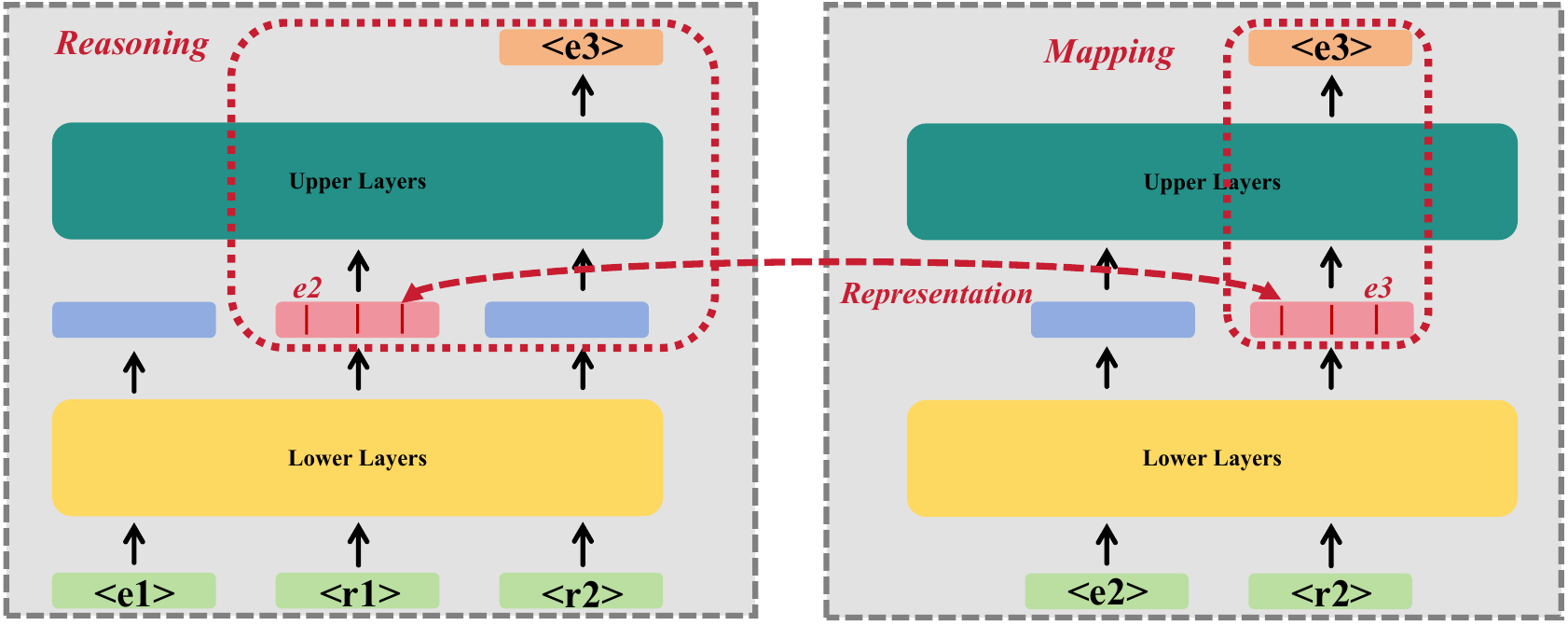}
\caption{\textbf{(Left)} In a two-hop query $(e_1,r_1,r_2)\!\rightarrow\!e_3$, lower layers first form the bridge entity representation $e_2$ at $r_1$. Upper layers must then combine $\mathscr{R}_{e_2}$ with $r_2$ to infer $e_3$. \textbf{(Right)} For a single-hop fact $(e_2,r_2,e_3)$, the representation of $e_3$ is already formed at $r_2$ in lower layers, so upper layers only map it to the output $h^L_{e_3}$ without relational reasoning.}

\label{fig:reasoning-mapping}

\end{figure*}

We next investigate why generalization emerges on Test-II and Test-OI, but consistently fails on Test-IO and Test-OO. An observation is that the essential difference between these test sets lies in whether the second hop corresponds to in-distribution or out-of-distribution knowledge. Given that the first-hop representation has already been shown to be well-formed and robust (Section \ref{section:4.2}), the remaining bottleneck must reside in the second-hop computation, namely, how the model operates on intermediate representations.

In a two-hop query of the form $(e_1, r_1, r_2) \rightarrow e_3$, during the second-hop inference, the model is required to use the representation of the bridge entity $e_2$ formed at position $r_1$, and then reason through relation $r_2$ to infer the target entity $e_3$. As shown on the left side of Figure \ref{fig:reasoning-mapping}, this requires the upper layers to implement a computation that implements $\mathscr{R}_{e_2} \xrightarrow{r_2} {e_3}$, where the input is an intermediate entity representation rather than a surface token embedding.

In contrast, during single-hop training on atomic facts $(e_2, r_2, e_3)$, the situation is fundamentally different. At position $r_2$, the upper layers already have direct access to the representation of $e_3$. Consequently, the upper layers \textbf{are not required} to perform relational reasoning; instead, they can learn a much simpler transformation: $\mathscr{R}_{e_3} \longrightarrow {e_3}$, namely a mapping from an already-formed representation to the final output space, as shown on the right side of Figure \ref{fig:reasoning-mapping}.

This distinction leads to a \textbf{training-inference mismatch} hypothesis: single-hop training primarily teaches the upper layers to perform representation mapping rather than representation-based reasoning. Under this view, the observed generalization behavior follows naturally. For Test-II and Test-OI, all their second hops in the form of intermediate representations have appeared during training  (i.e., within Train-II). Consequently, the model learns to consume the $\mathscr{R}_{e_2}$ and reason through $r_2$, enabling it to transfer this representation-based reasoning capability to these test subsets. In contrast, for Test-IO and Test-OO, the second hop has never been trained in a representational form—it has only been encountered as a single-hop atomic fact. The upper layers, therefore, attempt to apply a learned mapping where reasoning is required, leading to systematic failure.

To validate this hypothesis, we design a linear probing experiment\footnote{Besides,  we also conduct an attention blocking experiment \citep{geva-etal-2023-dissecting} in Appendix \ref{appendix:attention_mask}, which yields the same conclusion.}. Specifically, for atomic training examples, we collect: the hidden state at the $\langle r \rangle$ position from an intermediate layer $\ell$ as input, denoted $\mathbf{h}^{\ell}_{r}$; and the hidden state at the same position from the final layer $L$ as output, denoted $\mathbf{h}^{L}_{r}$. We then fit a linear model $\mathbf{h}^{L}_{r} \approx \mathbf{W}\mathbf{h}^{\ell}_{r}$ on a subset of these $(\mathbf{h}^{\ell}, \mathbf{h}^{L})$ pairs, and evaluate its performance on held-out data using both mean squared error (MSE) and cosine similarity as comprehensive evaluation metrics, similar to the idea in \citet{khandelwal2025language}. We compare the linearity of this mapping in two scenarios: the $r$ position for single-hop queries and the $r_2$ position for two-hop queries.

Figure \ref{fig:linear_prediction} reveals a clear divergence in linearity, particularly in layer 5, the bridge representation layer. \textbf{For single-hop queries}, the transformation from layer 5 to the final output exhibits near-perfect linearity (high cosine similarity, low MSE). This confirms our hypothesis that for atomic facts, once the entity is resolved, the upper layers function primarily as linear mappers. In contrast, \textbf{for two-hop queries}, the linearity at layer 5 is significantly lower. This disparity indicates that at this critical stage, a simple linear mapping is insufficient; instead, the model requires non-linear operations (e.g., attention mechanisms) to aggregate the bridge entity information from $r_1$ into $r_2$. The gradual improvement in linearity across subsequent layers reflects the progressive completion of this information aggregation.

Consequently, the failure of OOD generalization can be attributed to a functional mismatch: the upper layers are trained on atomic facts to perform linear mapping, but multi-hop reasoning requires them to execute non-linear information aggregation at the bridge layer.

\section{Bridging the Representation-Reasoning Gap for Robust Two-Hop Generalization}
\label{section:5}
\begin{figure}[t]
    \centering
    \begin{subfigure}[b]{\linewidth} 
        \centering
        \includegraphics[width=\linewidth]{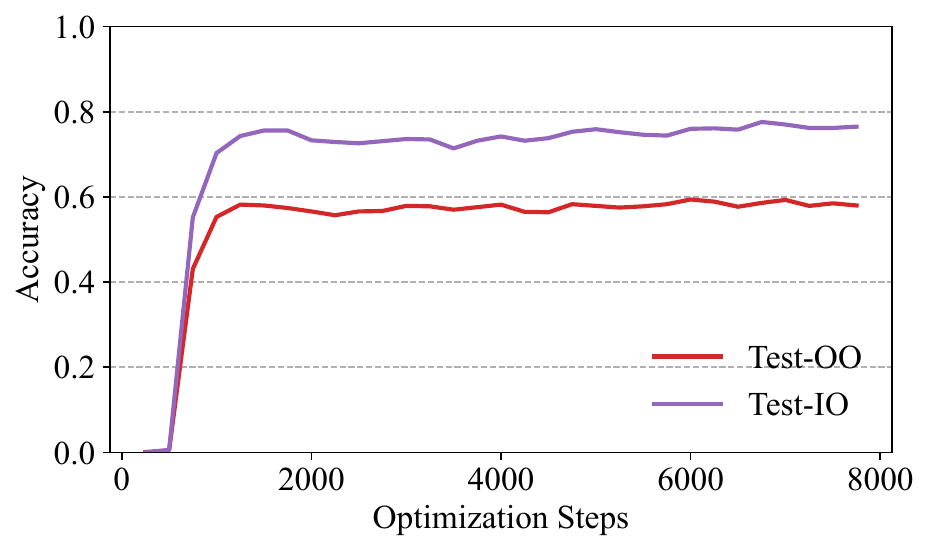}
        \caption{}
        \label{fig:rep-train}
    \end{subfigure}

    \begin{subfigure}[b]{\linewidth}
        \centering
        \includegraphics[width=\linewidth]{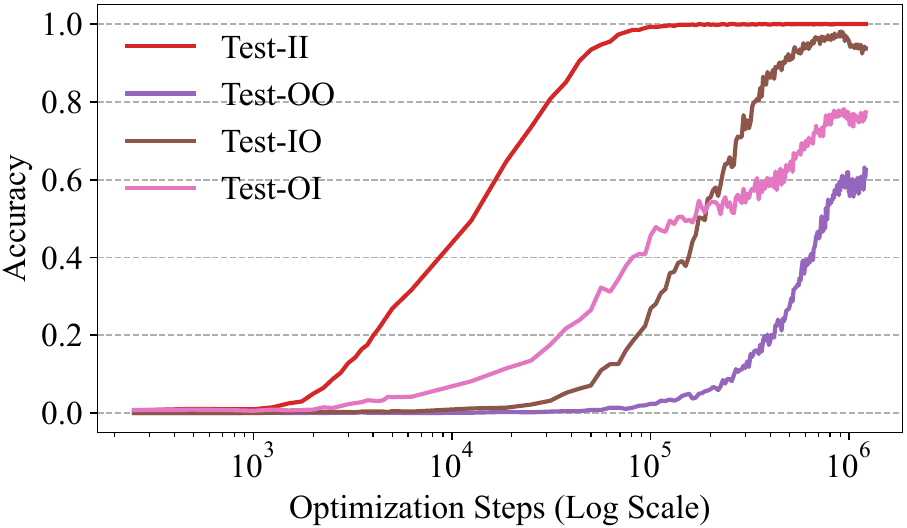}
        \caption{}
        \label{fig:loop-train}
    \end{subfigure}

    \caption{\textbf{(a)} Continual training of upper layers on atomic facts presented in a representational form. We only depict the process of continual training; the standard training process is the same as in Figure \ref{fig:training_results}. \textbf{(b)} Looped training with shared parameters yields consistent generalization across all test splits by aligning intermediate representations with lower-layer inputs.}
    \label{fig:solution}
\end{figure}
Our analysis so far indicates that the failure of two-hop generalization does not stem from missing knowledge, but is most likely caused by a structural mismatch between the roles played by different layers of the model. This diagnosis suggests a natural direction for achieving robust generalization: bridging the gap between lower-layer representations and upper-layer reasoning. In the following, we explore two strategies to achieve this goal.

\subsection{Explicitly training upper layers for representation-based reasoning}
\label{section:5.1}

A direct approach is to equip the upper layers with the ability to reason over representations. Concretely, after standard training, we introduce a continuous training process applied only to the upper layers (layer 5 and above). In this auxiliary phase, the model is trained on OOD atomic facts, but presented purely in a representational form: the inputs to the upper layers are pairs of hidden states $(h_e, h_r)$, rather than surface tokens (It is necessary to note that this continual training only exposes the model to atomic facts and does not leak any two-hop supervision). We provide more details about the continual training process in Appendix \ref{appendix:rep-based-training}.

The evaluation results on Test-OO and Test-IO as training processes are shown in Figure \ref{fig:rep-train}. We observe that this additional process substantially improves generalization on previously failing settings, namely Test-IO and Test-OO. This improvement confirms our earlier diagnosis that the upper layer struggles to reason over out-of-distribution atomic facts in representational form. However, this approach comes with notable drawbacks: it introduces additional training stages and increases computational cost.

\subsection{Eliminating layer-wise mismatch via looped architecture}
\label{section:5.2}

An alternative and more principled solution is to remove the structural mismatch altogether. Instead of explicitly training the upper layers on representational inputs, we enforce alignment between representations and lower-layer inputs by sharing parameters across layers (Here, we share the parameters of the top four layers and the bottom four layers), forming a \textbf{looped architecture} \citep{yang2024looped, fan2025looped}. In this design, the hidden states produced by the model's first forward pass are fed back as input to a second forward pass using the same set of parameters. Because both passes share weights, the model is forced to process surface tokens and intermediate representations in a unified format.

As a result, even when the model has never encountered OOD atomic knowledge in a representational form, the alignment between representations and lower-layer inputs allows it to reuse the same reasoning circuit, thus bridging the training-inference gap. Figure \ref{fig:loop-train} demonstrates that this looped training strategy yields substantial improvement in generalization performance, without requiring additional supervision on OOD representations.

\vspace{-2mm}
\subsection{Evidence of Representation-Input Alignment under Looped Training}
\label{section:5.3}

The intuition behind looped training is that robust two-hop generalization emerges from an explicit alignment between intermediate representations and lower-layer inputs. In this subsection, we provide direct empirical evidence for this claim by measuring representational similarity across layers.

To quantify the alignment between the model's internal representations and its input forms, we calculate the cosine similarity for both the bridge entity $e_2$ and the relation $r_2$ at critical transition layers. We define two alignment metrics:
\begin{enumerate}
    \item \textbf{Loop-End Alignment ($S_{\text{end}}$)}: Measures how close the representation at the end of the loop block (layer 4) is to the input embeddings. $S_{\text{end}}(e_2) = \cos(\mathbf{h}^4_{r_1}, \mathbf{E}_{e_2}), \: S_{\text{end}}(r_2) = \cos(\mathbf{h}^4_{r_2}, \mathbf{E}_{r_2})$

    \item \textbf{Bridge-Layer Alignment ($S_{\text{bridge}}$)}: Measures the similarity between the bridge layer (layer 5) and layer 1. $S_{\text{bridge}}(e_2) = \cos(\mathbf{h}^5_{r_1}, \mathbf{h}^1_{e_2}), \: S_{\text{bridge}}(r_2) = \cos(\mathbf{h}^5_{r_2}, \mathbf{h}^1_{r_2})$
\end{enumerate}
\begin{table}[t]
\centering
\small
\setlength{\tabcolsep}{5pt} 
\begin{tabular}{lcccc}
\toprule
\multirow{2}{*}{\textbf{Training Strategy}} & \multicolumn{2}{c}{\textbf{$S_{\text{end}}$}} & \multicolumn{2}{c}{\textbf{$S_{\text{bridge}}$}} \\
\cmidrule(lr){2-3} \cmidrule(lr){4-5}
& $e_2$ & $r_2$ & $e_2$ & $r_2$ \\
\midrule
Standard & 0.3528 & 0.4121 & -0.0028 & 0.3052 \\
\textbf{Looped } & \textbf{0.6796} & 0.3141 & \textbf{0.8132} & \textbf{0.8487} \\
\bottomrule
\end{tabular}
\caption{Cosine similarity analysis of representation alignment. Looped training forces high compatibility between intermediate representations and lower-layer inputs.}
\vspace{-3mm}
\label{tab:alignment}
\end{table}

The results in Table \ref{tab:alignment} reveal a striking contrast between the two training regimes. 
Under looped training, intermediate representations exhibit strong alignment with lower-layer inputs. 
In particular, the representation of the bridge entity $e_2$ shows consistently high cosine similarity across layers, and the representation of relation $r_2$ becomes highly aligned when comparing higher-layer outputs to lower-layer inputs. 
In contrast, under standard training, this alignment is weak or absent.

Driven by the need for parameter sharing (e.g., layer 5 and layer 1 have the same parameters), the model is compelled to converge their hidden states into a unified format. This alignment ensures the reasoning machinery learned on surface forms can be seamlessly re-applied to intermediate states, explaining the generalization observed in Figure \ref{fig:loop-train}.

\subsection{Scalability of Looped Architecture}

We finally examine whether the benefits of looped training extend to modern architectures and real-world two-hop data. 
We benchmark performance on two model backbones: the GPT-style model used in previous sections and a Llama-based architecture. 
We also move beyond symbolic tasks to a real natural language dataset derived from Wikidata5M, consisting of natural-language multi-hop reasoning samples \footnote{The construction details are shown in Appendix \ref{appendix:real_two_hop}.}. 
Results are reported in Table \ref{tab:scalability}. 
Loops prove robust on modern architecture. On the real-world dataset, the model successfully generalizes to OOD splits, confirming that the convergence of intermediate states is not an artifact of symbolic data but a viable strategy for realistic reasoning tasks. 
We also study how changing the number of trainable layers under the looped architecture affects generalization; results in Appendix \ref{appendix:scale} show that allowing more trainable layers substantially improves the model’s generalization ability and breaks the bottleneck in Figure \ref{fig:solution}.

\begin{table}[h]
    \centering
    \small
    \setlength{\tabcolsep}{4pt} 
    \begin{tabular}{l cccc}
    \toprule
    \textbf{Setup} & \textbf{Test-II} & \textbf{Test-IO} & \textbf{Test-OO} & \textbf{Test-OI} \\
    \midrule
    GPT-Symbolic & 1.0 & 0.97 & 0.84 & 0.97 \\
    Llama-Symbolic & 1.0 & 0.83 & 0.54 & 0.99 \\
    \midrule 
    GPT-Real & 1.0 & 0.79 & 0.66 & 0.98 \\
    Llama-Real & 0.99 & 0.74 & 0.62 & 0.98 \\
    \bottomrule
    \end{tabular}
    \caption{Generalization performance across different architectures and datasets. \textit{Real} denotes natural language 2-hop data extracted from Wikidata5M.}
    \vspace{-5mm}
\label{tab:scalability}
\end{table}

\section{Related Works}

\subsection{Mechanistic Interpretability}

Mechanistic interpretability seeks to reverse engineer how LLMs implement specific computations \citep{olah2022mechinterp}. A line of work studies hidden states across layers, including logit lens methods that project intermediate states through the unembedding matrix to examine token-level predictions \citep{nostalgebraist2020logitlens, dar-etal-2023-analyzing, yu-ananiadou-2024-interpreting, katz-belinkov-2023-visit}. Other approaches use causal interventions on hidden states to measure their effect on outputs and localize functionally relevant components \citep{stolfo-etal-2023-mechanistic, meng2022locating, Vig2020InvestigatingGB}. Motivated by the superposition hypothesis, sparse autoencoders extract more interpretable features from activations \citep{scherlis2022polysemanticity, elhage2022toy, balcells2024evolution, zhao-etal-2025-steering}. Together, these studies provide a toolkit for dissecting how complex behaviors emerge from layered computations.

\subsection{Implicit Reasoning}
LLMs can perform multi-step or compositional reasoning without explicitly generating intermediate chains, a phenomenon termed implicit reasoning \citep{kojima2022large, wei2022chain}. Prior work shows that transformers encode latent reasoning states internally \citep{deng2024explicit}, and pretrained models can answer multi-hop queries without exposing intermediate steps \citep{yang2024large, balesni2024lessons}. Mechanistic analyses further link the emergence of implicit reasoning to representation structure and grokking-like dynamics \citep{wang2024grokking, ye2026transformers}. Based on this line of work, we attribute second-hop OOD failure to a layer-wise mismatch between representation and reasoning.

\subsection{Two-Hop Reasoning}

Two-hop reasoning is a canonical multi-hop QA setting that requires retrieving and composing multiple atomic facts into a single inference \citep{yang-etal-2018-hotpotqa}. Prior work identifies a persistent gap between factual recall and compositional reasoning in LLMs: \citet{press-etal-2023-measuring} formalizes this as the compositionality gap, where models fail on two-hop queries despite succeeding on the corresponding single-hop facts. Subsequent studies show that such failures are unstable across domains and settings, questioning whether LLMs robustly perform genuine multi-hop reasoning \citep{berglund2023taken, yang2024large}. Mechanistic analyses further reveal that even when the bridge entity is correctly retrieved at the first hop, it is often not utilized by higher layers, indicating a disconnect between representation formation and reasoning \citep{biran-etal-2024-hopping, yu2025back}. Related work also studies two-hop reasoning circuits from theoretical, controlled finetuning, and in-context learning perspectives \citep{pmlr-v267-feng25m, guo2025llms}.

\section{Conclusion}

This work provides a mechanistic explanation of the two-hop problem in Transformers. In a controlled symbolic setting, we identify an asymmetric pattern: models generalize when the second hop is in-distribution but fail when it is out-of-distribution. We trace this behavior to representation dynamics: successful cases exhibit consistent bridge-entity representations across contexts, whereas failures arise because upper layers mainly map resolved representations to outputs rather than reason over intermediate states. Based on these findings, we propose a looped architecture that aligns intermediate representations with lower-layer inputs, enabling reuse of the same reasoning circuitry across input forms. Experiments on both symbolic and natural-language tasks show that the looped architecture improves OOD generalization.


\section*{Limitations}

While our looped architecture effectively enables representation reuse and substantially improves generalization, our training procedure remains relatively simple: we adopt a standard pre-training strategy and apply only a single loop. It is possible that alternative training schedules, multiple looping iterations, or more adaptive mechanisms could further enhance efficiency or performance \citep{zhu2025scaling}. Exploring these directions may lead to more effective architectures, but it falls outside the scope of the present study.

\section*{Acknowledgments}
Zili Zhang ran all the experiments in the paper and drafted part of the paper. Yilin Wang designed the conceptual framework, designed the experiments with Zili Zhang, and drafted part of the paper. Heng Wang and Herun Wan provided constructive feedback on the paper draft. Minnan Luo provided feedback throughout the project. Besides, we also thank all LUD Lab members for our collaborative research environment. 

This work is supported by the Fundamental and Interdisciplinary Disciplines Breakthrough Plan of the Ministry of Education of China (No. JYB2025XDXM101), the National Natural Science Foundation of China (No. 62272374), the Natural Science Foundation of Shaanxi Province (No. 2024JC-JCQN-62), Huawei-Xi'an Jiaotong University Elite Class Program (No. INCCHN2508012229), the State Key Laboratory of Communication Content Cognition under Grant No. A202502, and the Key Research and Development Project in Shaanxi Province (No. 2023GXLH-024).

\bibliography{main}

\appendix
\clearpage

\section{Dataset Details}
\label{appendix:dataset-details}
Our training set is constructed following the approach of \citet{wang2024grokking}, while the test set is built following the methodology of \citet{ye2026transformers}. 
Specifically, we first construct 2,000 distinct entities (\texttt{<e\_id>}) and 200 distinct relations (\texttt{<r\_id>}).

For atomic knowledge construction, each entity is randomly assigned 20 distinct outgoing relations, where each relation points to another entity—for example, $(\langle e_0 \rangle, \langle r_0 \rangle) \rightarrow \langle e_1 \rangle$. This results in a total of 40,000 atomic facts. To create in-distribution (ID) and out-of-distribution (OOD) facts, 95\% of the atomic facts are assigned to the in-distribution set (ID\_atomic), and the remaining 5\% are assigned to the out-of-distribution set (OOD\_atomic).

Two atomic facts can be combined into a two-hop fact if the tail entity of one fact matches the head entity of the other. 
In a two-hop fact, the first-hop and second-hop can each come from either the ID or the OOD atomic set.
We first consider two-hop facts where both the first and second hops are drawn from the ID atomic set.
These two-hop facts are further divided: one portion is included in the test set as Test-II, and from the remaining portion, a number of two-hop facts corresponding to $\phi$ times the number of ID atomic facts are randomly selected and included in the training set.

Additionally, we construct Test-IO, Test-OI, and Test-OO to more comprehensively evaluate the model's generalization ability. Specifically, Test-IO contains two-hop facts where the first hop comes from ID atomic facts and the second hop comes from OOD atomic facts; Test-OI and Test-OO are defined analogously.

In this paper, we set $\phi=7.2$ because, according to \citet{wang2024grokking}, 7.2 represents a relatively intermediate value. This corresponds to a training set size that allows the model to generalize on Test-II while avoiding an excessively large dataset that would significantly increase training cost. Choosing $\phi=7.2$ is appropriate for our study of why the model fails to generalize on Test-IO and Test-OO.

More detailed information about the dataset is provided in Table~\ref{tab:dataset-format} and Table~\ref{tab:dataset-stats}. Table~\ref{tab:dataset-format} shows the formats of atomic and two-hop facts. Table~\ref{tab:dataset-stats} presents the number of facts in each data split.

\begin{table}[H]  
\centering
\begin{tabular}{c c}  
\hline
Fact Type & Example \\
\hline
Atomic Fact & \texttt{<e\_0><r\_0><e\_1>} \\
Two-hop Fact & \texttt{<e\_0><r\_0><r\_1><e\_2>} \\
\hline
\end{tabular}
\caption{Example formats of facts in the dataset.}
\label{tab:dataset-format}
\end{table}

\begin{table}[H]
\centering
\begin{tabular}{c c c}  
\hline
Source & Data Type & Quantity \\
\hline
\multirow{3}{*}{Training Set} & ID\_atomic & 38000 \\
                               & OOD\_atomic & 2000 \\
                               & Train-II    & 273600 \\
\hline
\multirow{7}{*}{Test Set}     & ID\_atomic  & 3000 \\
                               & OOD\_atomic & 2000 \\
                               & Train-II    & 3000 \\
                               & Test-II     & 3000 \\
                               & Test-IO     & 3000 \\
                               & Test-OI     & 3000 \\
                               & Test-OO     & 1987 \\
\hline
\end{tabular}
\caption{Detailed statistics of the dataset.}
\label{tab:dataset-stats}
\end{table}

\section{Training Details}
\label{appendix:exp-env}
All experiments are conducted on Huawei Ascend 910B NPUs. Training is performed using four Huawei Ascend 910B (64GB) NPUs, with a batch size of 1024 per NPU. We use a learning rate of $1 \times 10^{-4}$ and a weight decay of 0.1 for optimization. Inference on the test set is conducted using greedy decoding. The number of optimization steps and corresponding training time for the main training experiments reported in this paper are summarized in Table~\ref{tab:training-details}.


\begin{table}[H]
\centering
\small
\resizebox{\columnwidth}{!}{%
\begin{tabular}{c c c}
\toprule
Experiment & Steps & Training Time (h) \\
\midrule
Standard Training & 3,500,000 & 263.05 \\
Rep.-Based Training & 7,750 & 0.88 \\
Looped Training & 1,212,500 & 86.80 \\
\bottomrule
\end{tabular}
}
\caption{Optimization steps and training time for the main training experiments in this paper. 
Standard Training corresponds to Section~\ref{section:3}, Rep.-Based Training (denoting representation-based training) corresponds to Section~\ref{section:5.1}, and Looped Training corresponds to Section~\ref{section:5.2}.}
\label{tab:training-details}
\end{table}

\section{Intermediate Representations of Bridge Entities Reflect Graph Structure}
\label{appendix:bridge-graph-structure}

In this section, we explore the formation of entity representations in relation to the graph structure within the dataset. We first analyze the structure of the graph composed of all atomic facts, and then examine how this graph structure influences the formation of entity representations.

\begin{figure*}[t]
    \centering
    \begin{subfigure}{0.45\textwidth}
        \centering
        \includegraphics[width=\linewidth]{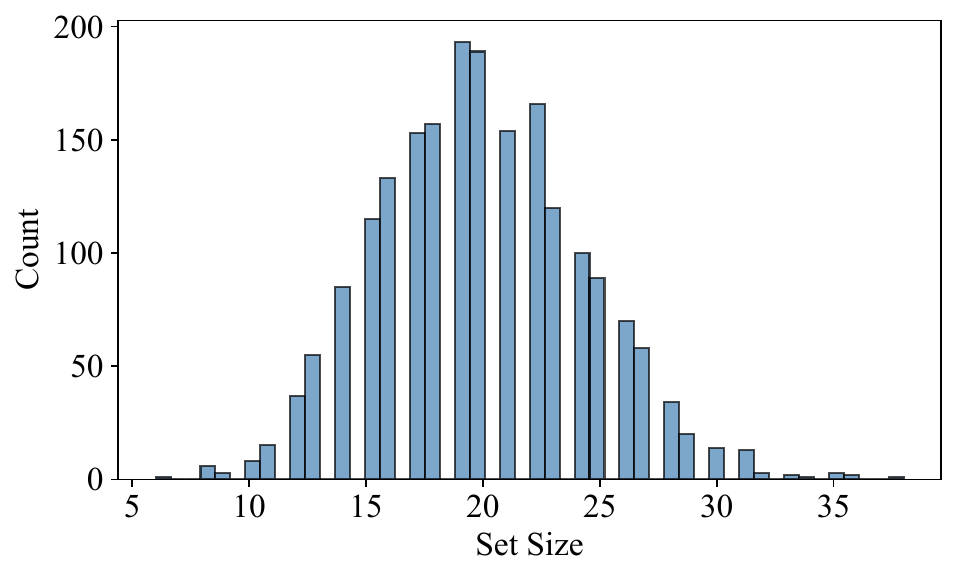}
        \caption{Source entity set $e_1$}
        \label{fig:e2e1_hist}
    \end{subfigure}
    \begin{subfigure}{0.45\textwidth}
        \centering
        \includegraphics[width=\linewidth]{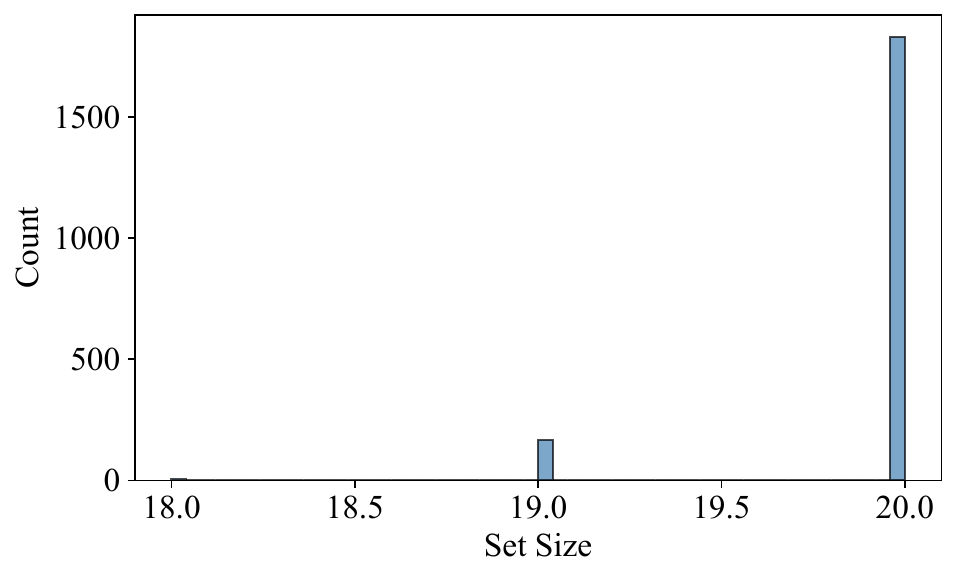}
        \caption{Target entity set $e_3$}
        \label{fig:e2e3_hist}
    \end{subfigure}

    \begin{subfigure}{0.45\textwidth}
        \centering
        \includegraphics[width=\linewidth]{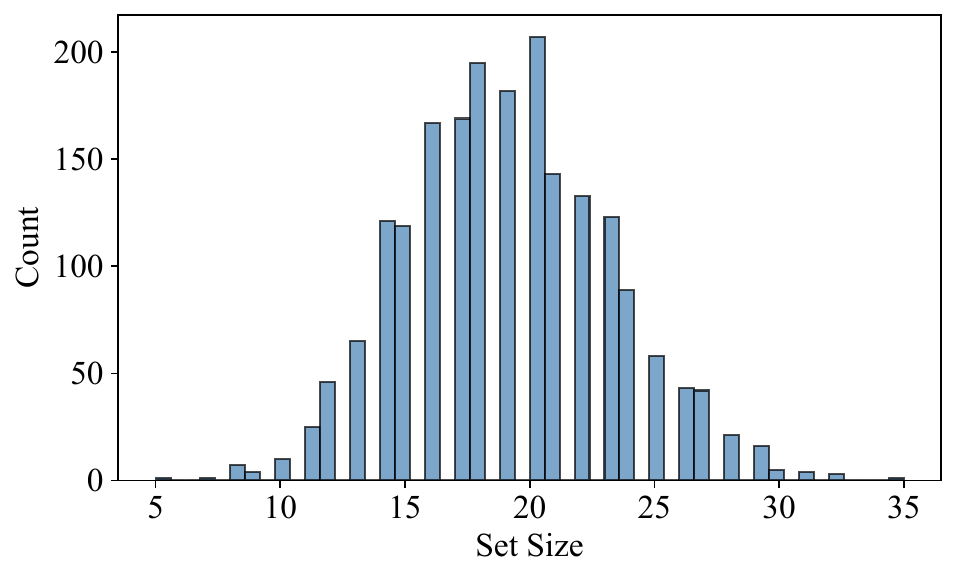}
        \caption{Incoming relation set $r_1$}
        \label{fig:e2r1_hist}
    \end{subfigure}
    \begin{subfigure}{0.45\textwidth}
        \centering
        \includegraphics[width=\linewidth]{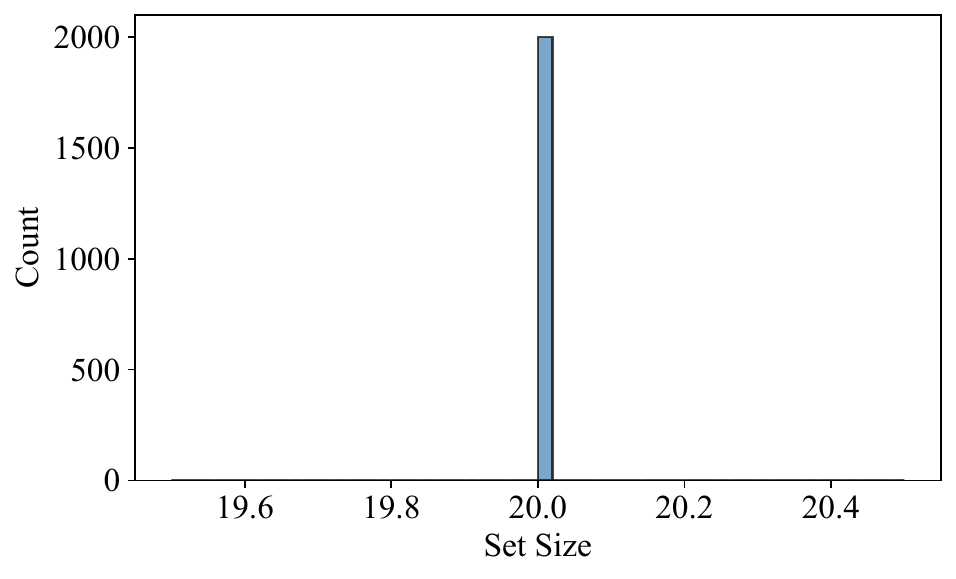}
        \caption{Outgoing relation set $r_2$}
        \label{fig:e2r2_hist}
    \end{subfigure}

    \caption{Histograms of the set size distributions for the $e_1$, $e_3$, $r_1$, and $r_2$ sets across all entities.}
    \label{fig:set_size_hists}
\end{figure*}

\begin{figure}[t]
    \centering
    \includegraphics[width=\linewidth]{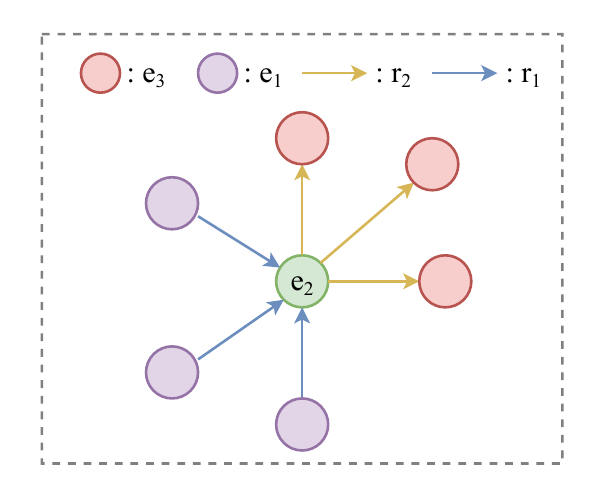}
    \caption{Illustration of the local graph structure centered at entity $e_2$, where $r_1$ and $r_2$ denote the sets of incoming and outgoing relations, respectively, connecting source entities $e_1$ and target entities $e_3$.}
    
    \label{fig:graph_structure_def}
\end{figure}

\subsection{Graph Structure in the Dataset}
\label{appendix:dataset-graph-structure}

The 40,000 atomic facts constructed in the dataset jointly form a graph structure.
The resulting graph contains 2,000 nodes, where each node corresponds to an entity. The graph includes 200 edge types, with each edge type representing a distinct relation. Each node emits 20 outgoing edges connecting to other nodes, and every node is also the target of incoming edges from other nodes.

For ease of presentation, we consider an entity $e_2$. 
The edges emitted by $e_2$ define a set of outgoing relations $r_2$, and the target entities of $r_2$ form a set denoted as $e_3$. 
The edges pointing to $e_2$ define a set of incoming relations $r_1$, and the source entities of $r_1$ form a set denoted as $e_1$. 
Consequently, each entity is associated with four sets: a set of target entities $e_3$, a set of outgoing relations $r_2$, a set of source entities $e_1$, and a set of incoming relations $r_1$, as illustrated in Figure~\ref{fig:graph_structure_def}.

For all entities, we collect statistics on the sizes of the $e_1$, $r_1$, $e_3$, and $r_2$ sets, and plot histograms of the size distributions for each of the four sets, as shown in Figure~\ref{fig:set_size_hists}. As can be clearly observed, the sizes of the $e_1$ and $r_1$ sets approximately follow a Gaussian-like distribution centered around 20. In contrast, the size of the $r_2$ set is exactly 20 for all entities, which is consistent with our atomic fact construction procedure. The size of the $e_3$ set is 20 for the majority of entities, with a small fraction having size 19. This deviation arises because, among the 20 outgoing edges of an entity, multiple edges may point to the same target entity, resulting in a reduced number of unique entities in $e_3$.

\subsection{How Does Graph Structure Shape the Formation of Entity Representations?}
\label{appendix:graph-structure-entity-representations}

\begin{figure*}[t]  
    \centering
    \begin{subfigure}[b]{0.45\linewidth}
        \centering
        \includegraphics[width=\linewidth]{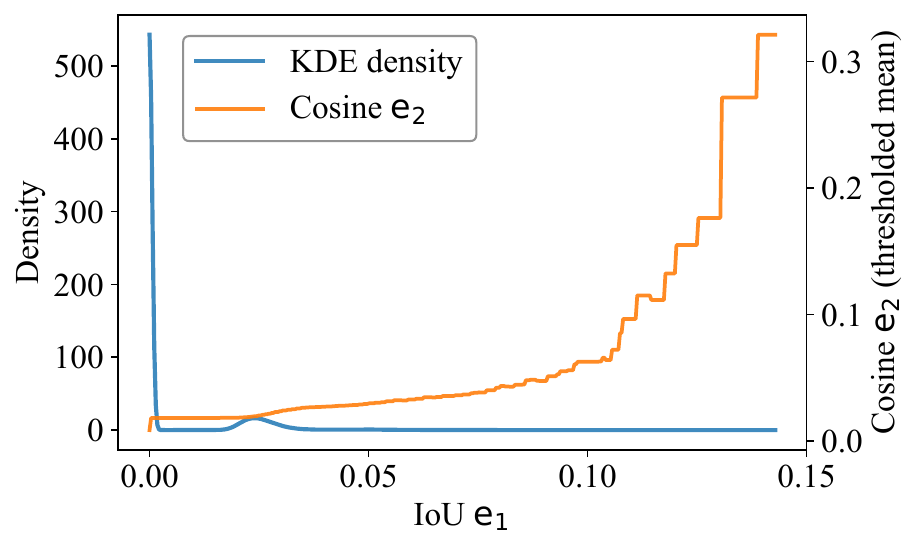}
    \end{subfigure}
    \hfill
    \begin{subfigure}[b]{0.45\linewidth}
        \centering
        \includegraphics[width=\linewidth]{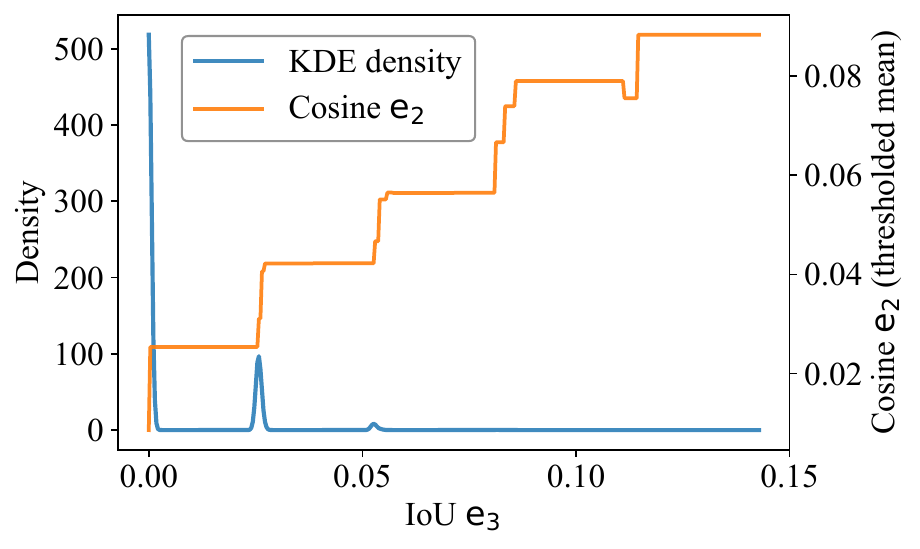}
    \end{subfigure}

    \vspace{2mm}

    \begin{subfigure}[b]{0.45\linewidth}
        \centering
        \includegraphics[width=\linewidth]{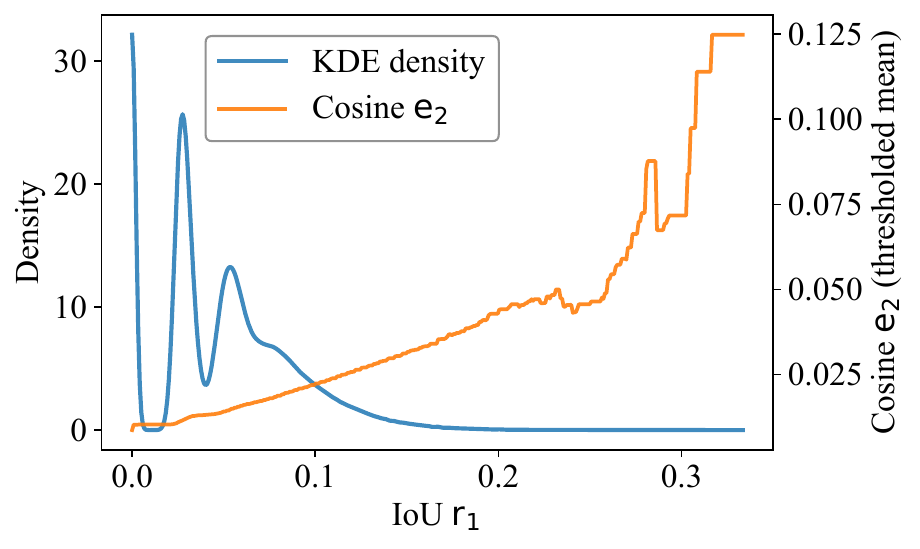}
    \end{subfigure}
    \hfill
    \begin{subfigure}[b]{0.45\linewidth}
        \centering
        \includegraphics[width=\linewidth]{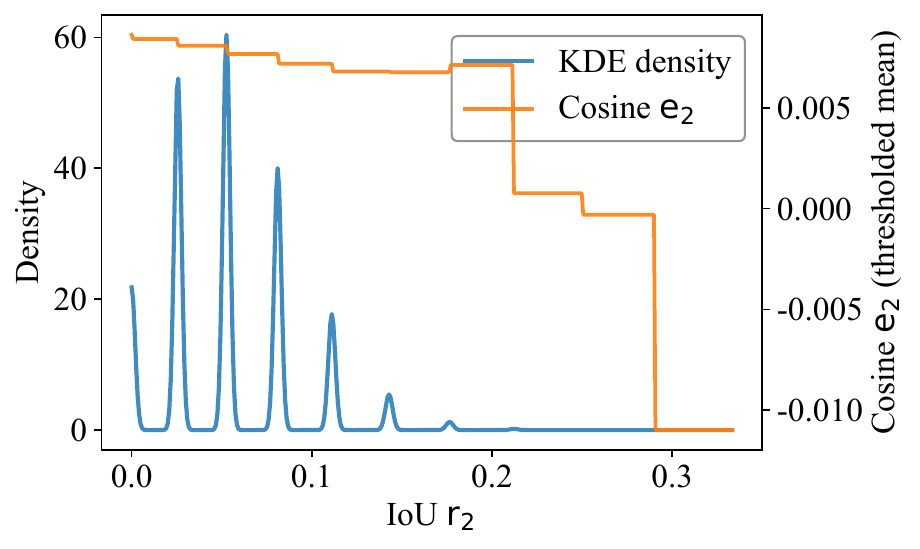}
    \end{subfigure}

    \caption{Relationship between pairwise set overlap and representation similarity for the four attribute sets $e_1$, $e_3$, $r_1$, and $r_2$. The x-axis denotes the intersection-over-union (IoU) between the corresponding sets of two entities. The left y-axis (blue curves) shows the kernel density estimation (KDE) of entity pairs at each IoU value. The right y-axis (orange curves) reports the average cosine similarity of the intermediate representations for entity pairs with IoU greater than or equal to the corresponding threshold.}
    \label{fig:iou_kde_thresholds}
\end{figure*}

In this paper, we identify and define the output hidden state at $r_1$ in layer 5 as the intermediate representation of the bridge entity. We further observe that, as training progresses, the intermediate representations of the same bridge entity in different contexts gradually converge. A natural question arises: what does this aggregated entity representation reflect, or what determines its position in the representation space?

An intuitive hypothesis is that the formation of entity representations is related to the graph structure within the dataset. Specifically, if two entities have similar surrounding attributes—i.e., high overlap in their $e_1$, $r_1$, $e_3$, and $r_2$ sets—then their representations in the representation space should also be close.

Based on this hypothesis, we design the following experiment to empirically examine the relationship between local graph structure and entity representation similarity:

\begin{enumerate}
    \item For each entity, extract its surrounding attribute set, consisting of $e_1$, $r_1$, $e_3$, and $r_2$.
    \item Collect the intermediate hidden state representations of all entities using ID atomic facts, and compute the mean representation for each entity.
    \item Pair all entities and compute the overlap between their corresponding attribute sets, and plot kernel density estimation (KDE) curves for the resulting overlap distributions.
    \item For multiple overlap thresholds, calculate the average cosine similarity between the representations of all entity pairs exceeding each threshold.
    \item Plot the average cosine similarity as a function of the overlap threshold.
\end{enumerate}

This experiment enables us to investigate how local graph structure influences the formation of entity representations.
The results are shown in Figure~\ref{fig:iou_kde_thresholds}.
The following provides a detailed analysis of these results:

\begin{figure*}[t]
\centering

\begin{subfigure}{0.48\textwidth}
    \centering
    \includegraphics[width=\linewidth]{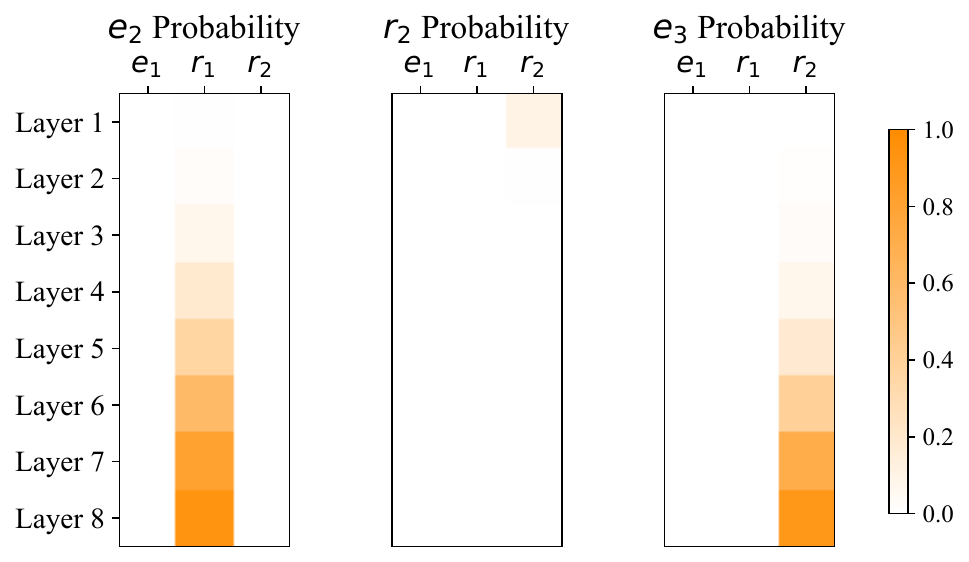}
    
    \vspace{2pt}
    {\small (a)}
\end{subfigure}
\hfill
\begin{subfigure}{0.48\textwidth}
    \centering
    \includegraphics[width=\linewidth]{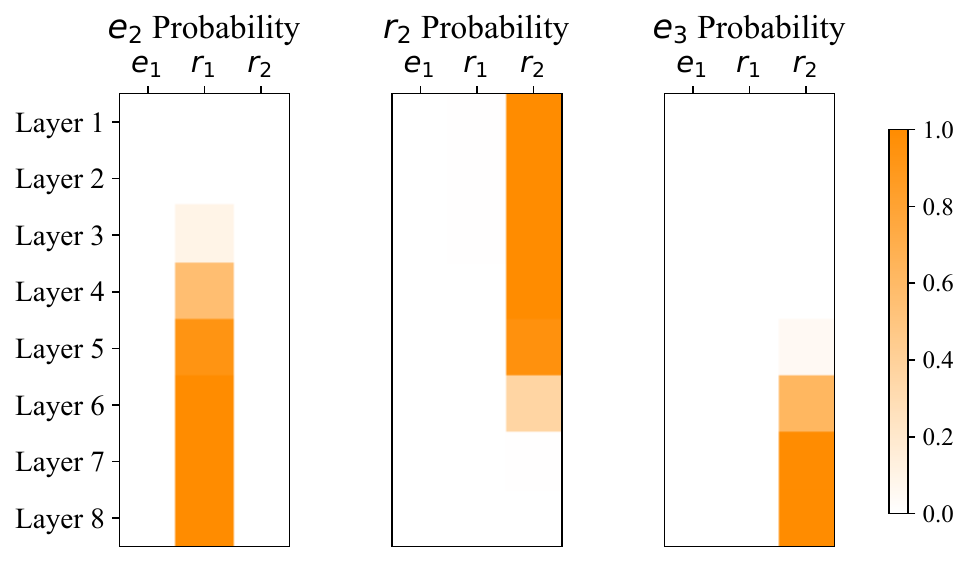}
    
    \vspace{2pt}
    {\small (b)}
\end{subfigure}

\caption{Logit lens results at different stages of the standard training process. 
(a) Memorization stage (checkpoint 3k). 
(b) First generalization stage (Test-II generalization, checkpoint 100k).}
\label{fig:logit_lens_training}

\end{figure*}

\vspace{0.5em}
\noindent\textbf{KDE distributions.}  
We first examine the KDE curves. Most entity pairs concentrate at low set overlap values, indicating that only a small fraction of entity pairs share highly similar surrounding attribute sets.

\vspace{0.5em}
\noindent\textbf{Cosine similarity trends.}  
Next, we consider the cosine similarity curves. For the $e_1$, $e_3$, and $r_1$ sets, the cosine similarity consistently increases as the overlap threshold grows, consistent with the expectation that greater structural overlap corresponds to more similar representations. In contrast, the $r_2$ set shows decreasing cosine similarity as the overlap threshold increases, appearing to contradict this expectation.

\vspace{0.5em}
\noindent\textbf{Magnitude and stability of effects.}  
A closer inspection reveals an important difference. The cosine similarity curves for $e_1$, $e_3$, and $r_1$ remain above zero and exhibit relatively large variations, whereas the $r_2$ curve is roughly centered around zero, with both positive and negative effects and a much smaller range of variation. This suggests that the overlap of the $r_2$ set has only a negligible influence on the cosine similarity between entity representations.

\vspace{0.5em}
Overall, higher overlap in the $e_1$, $e_3$, and $r_1$ attribute sets corresponds to greater similarity between the representations of two entities. This indicates that the intermediate entity representations capture implicit graph structures within the dataset, and that, as training progresses, the model tends to organize its hidden representations in a more structured and efficient manner.

\section{More Detailed Logit Lens Results}
\label{appendix:logit_lens_details}

Figure~\ref{fig:logit_lens} in Section~\ref{section:4.1} shows the logit lens results of the final model, where the bridge entity representation is identified as $h_{r_1}^5$ by locating the starting point of the second hop. As a supplement to that analysis, Figure~\ref{fig:logit_lens_training} presents logit lens results from earlier training stages.

During the memorization stage (Figure~\ref{fig:logit_lens_training}a), the probability of $e_3$ becomes significant even when the representation of the bridge entity $e_2$ is still not clearly formed. This indicates that, at this stage, the model does not rely on the bridge entity representation when answering two-hop queries, but rather memorizes the final two-hop answer directly, without performing genuine multi-hop reasoning.

In the first generalization stage (Figure~\ref{fig:logit_lens_training}b), the logit lens result is consistent with that of the final model. At the $r_2$ position, the probability of $r_2$ begins to decrease significantly at Layer 6, while the probability of $e_3$ starts to increase, marking the beginning of the second hop. This indicates that the bridge entity representation emerges at $h_{r_1}^5$, providing empirical support for the choice made in Section~\ref{section:4.2}, where this representation is patched at different stages of training.

\section{Layer-wise Attention Masking with a Sliding Window}
\label{appendix:attention_mask}

As discussed in Section~\ref{section:4.3}, when answering two-hop queries $(e_1,r_1,r_2,e_3)$, the upper layers (Layers~6, 7, and~8) of the model must perform a nonlinear transformation at the position of $r_2$, mapping $h_{e_2} \xrightarrow{r_2} e_3$. In contrast, when learning single-hop facts $(e_2,r_2,e_3)$, the upper layers only learn a linear mapping $h_{e_3} \rightarrow E_{e_3}$.
To further support the findings in Section~\ref{section:4.3}, we conduct a layer-wise attention masking experiment using a sliding window. The masking window spans three consecutive layers.

\begin{figure}[t]
    \centering
    \includegraphics[width=\columnwidth]{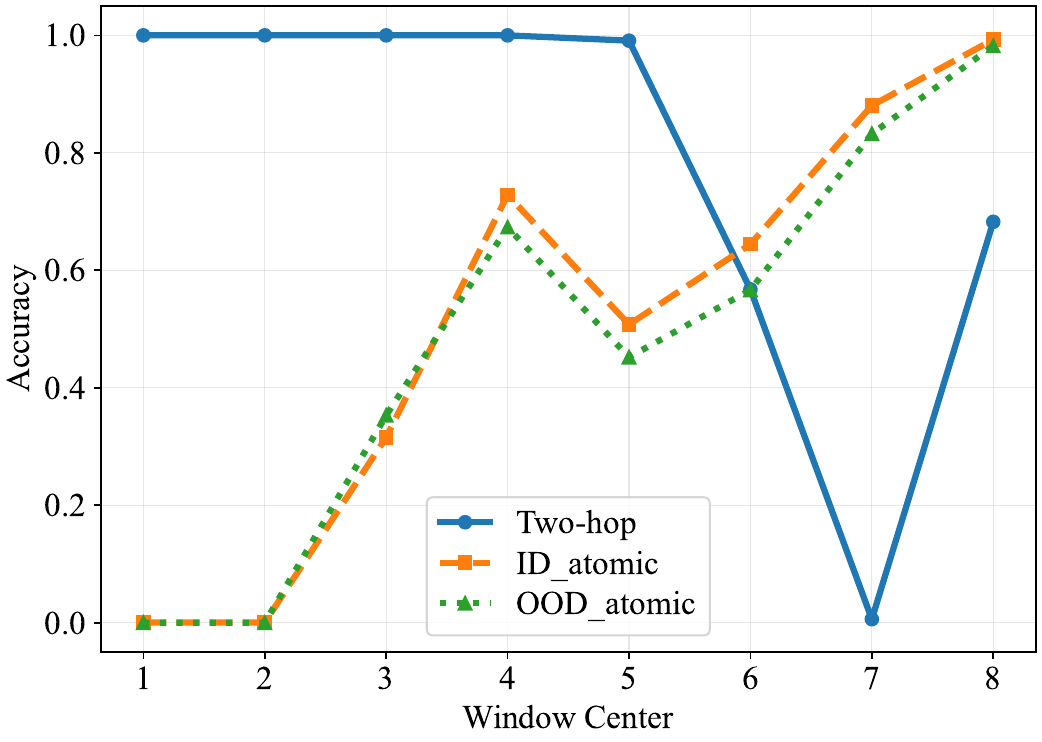}
    \caption{Accuracy of predicting the target entity under attention masking with a three-layer window. 
    For atomic facts $(e_2,r_2,e_3)$, we mask attention from $r_2$ to $e_2$; 
    for two-hop queries $(e_1,r_1,r_2,e_3)$, we mask attention from $r_2$ to $r_1$. 
    The x-axis denotes the center layer of the masking window.}
    \label{fig:attention_window}
\end{figure}

For atomic facts $(e_2, r_2, e_3)$, we mask the attention from $r_2$ to $e_2$ using a three-layer window with different window centers, and evaluate whether the model can still correctly predict $e_3$ after masking.  
For two-hop queries $(e_1, r_1, r_2, e_3)$, we instead mask the attention from $r_2$ to $r_1$ using the same window with different centers and test whether the model can still correctly predict $e_3$.
The purpose of this experiment is to verify whether the upper layers truly rely on information from the $e_2$ position when answering single-hop facts. The results are shown in Figure~\ref{fig:attention_window}.

As shown in Figure~\ref{fig:attention_window}, for atomic facts, masking attention in the lower layers significantly reduces the model's accuracy, whereas masking attention in the upper layers (i.e., when the masking window lies entirely within Layers~6--8) has only a minor effect.

In contrast, for two-hop queries, masking attention in the lower layers has almost no effect on the model's accuracy. However, when the masking window covers Layers~6--8, the model's accuracy drops to nearly zero.

These results confirm that when answering single-hop facts, the upper layers indeed do not require information from the $e_2$ position. Consequently, during training on atomic facts, the ability to perform single-hop reasoning is not learned in the upper layers. However, answering two-hop queries relies on this single-hop reasoning capability in the upper layers. This creates a mismatch between the capability required for two-hop reasoning and the capability actually learned from atomic facts during training. As a result, the model fails to generalize on Test-IO and Test-OO, since learning OOD atomic facts alone does not equip the upper layers with the ability to perform OOD single-hop reasoning.

\section{Details of Representation-based Training Experiments}
\label{appendix:rep-based-training}

\begin{figure}[t]   
    \centering
    \includegraphics[width=\linewidth]{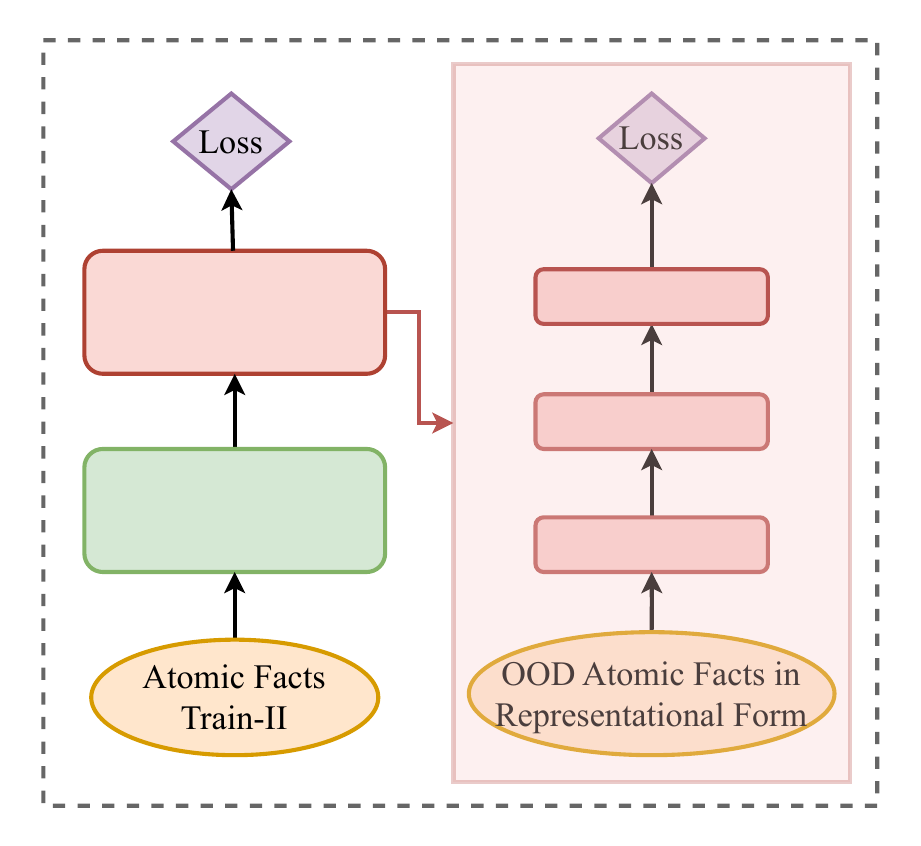}
    \caption{Illustration of the representation-based training process, where intermediate representations of entities and relations for OOD atomic facts are extracted from Train-II (layer 5) and used to train Layers 6--8.}
    \label{fig:rep_train_process}
\end{figure}

This section provides additional details on the training procedure described in Section~\ref{section:5.1}. The representation-based training procedure is illustrated in Figure~\ref{fig:rep_train_process}. In this experiment, besides the standard training procedure, we additionally train Layers 6--8 to learn OOD atomic facts $(e_2, r_2)$ in a representation-based form $(h_{e_2}, h_{r_2})$. The procedure is as follows:

\begin{enumerate}
    \item \textbf{Standard training:} We first train the model normally until it achieves high generalization performance on Test-II and Test-OI, as only when the model generalizes well on these benchmarks do the intermediate representations of the same entity become consistent and stable, which in turn provides a reliable foundation for subsequent representation-based training. For this experiment, we directly use the model after 3{,}500{,}000 optimization steps of standard training.

    \item \textbf{Multi-round training with representation-based reasoning:} The following procedure is repeated for multiple rounds. In each round, we first perform standard training to maintain the model's generalization performance and prevent the additional representation-based reasoning training from degrading it. After this, we collect the intermediate representations from Train-II: $h_e = h_5(r_1)$ for all entities and $h_r =h_5(r_2)$ for all relations. These representations are then used to construct representation-based forms $(h_{e_2}, h_{r_2})$ for all OOD atomic facts $(e_2, r_2)$. The pair $(h_{e_2}, h_{r_2})$ is fed into Layer 6 of the model and propagated through the subsequent layers, with supervision applied to guide the model to output the correct answers.
\end{enumerate}

Test accuracy is evaluated throughout this process.

\section{Logit Lens Analysis of Looped Training Models}
\label{appendix:looped-logit-lens}

\begin{figure}[t]
    \centering
    \includegraphics[width=\linewidth]{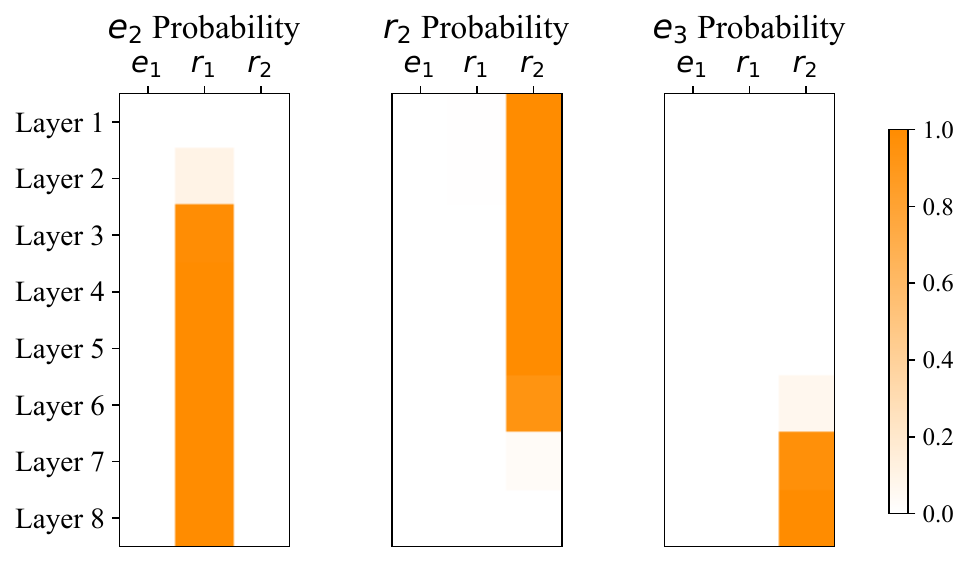} 
    \caption{Logit lens of the final model trained under the looped architecture, showing hidden state predictions for $e_2$, $e_3$, and $r_2$. Layer 6 marks the start of second-hop reasoning.}
    \label{fig:looped_logit_lens}
\end{figure}

Figure~\ref{fig:looped_logit_lens} shows the logit lens results of the final model trained under the recurrent architecture when answering two-hop questions. 
For a token at position \(t\) in layer \(l\), we extract its hidden state \(h_l(t)\), apply a LayerNorm, and project it through the model's embedding transpose \(E^\top\) to obtain the probability of each target entity or relation:
\begin{equation}
\begin{aligned}
p_l(x \mid t) &= \text{softmax}\Big(
E^\top \, \text{LayerNorm}(h_l(t))
\Big), \\
& \quad x \in \{e_2, e_3, r_2\}.
\end{aligned}
\label{eq:logit_lens}
\end{equation}
This allows us to inspect how the model’s predictions for \(e_2, e_3\) and \(r_2\) evolve across layers.

From Figure~\ref{fig:looped_logit_lens}, we observe that Layer 6 marks the onset of the second-hop reasoning: at the position of \(r_2\), the probability of \(r_2\) begins to decrease while the probability of \(e_3\) starts to increase. 
This indicates that Layer 6 is the first layer where the model initiates second-hop inference. 
Following the approach in Section~\ref{section:4.1}, we define the hidden state at layer 5 corresponding to the \(r_1\) position, denoted as \(h_5(r_1)\), as the intermediate representation of the bridge entity \(e_2\), and the hidden state at layer 5 corresponding to the \(r_2\) position, denoted as \(h_5(r_2)\), as the intermediate representation of relation \(r_2\). 
During the second-hop reasoning, the model performs inference based on these intermediate representations \(h_5(r_1)\) and \(h_5(r_2)\).

\section{Exploring Equivalent Alternatives to the Shared-Parameter Model}
\label{appendix:equivalent-alternatives}

\begin{figure*}[t]
  \centering
  \includegraphics[width=\textwidth]{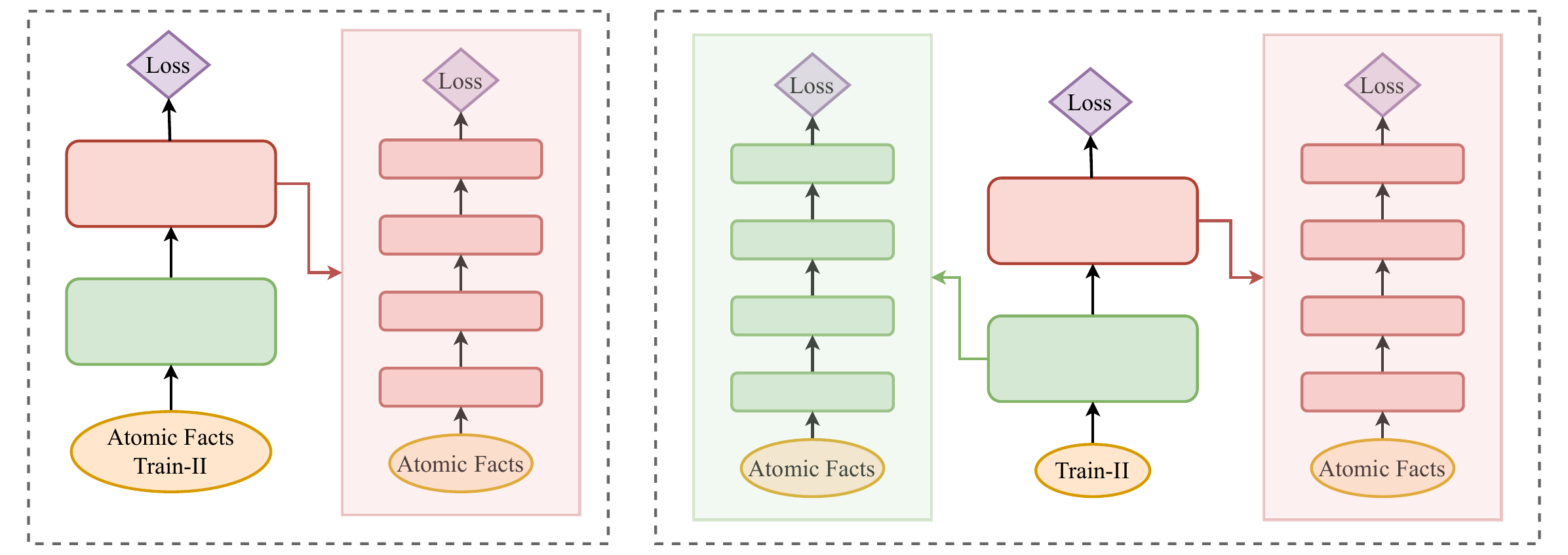}
  \caption{Comparison of two training strategies described in Appendix~\ref{appendix:equivalent-alternatives}.
  \textbf{Left:} the training strategy described in Appendix~\ref{appendix:equivalent-alternatives-upper}, where in each training round, embeddings of atomic facts are additionally used to train the upper four layers.
  \textbf{Right:} the training strategy described in Appendix~\ref{appendix:equivalent-alternatives-upper-lower}, where atomic facts are removed from the original training set, and in each training round, embeddings of atomic facts are additionally used to train both the upper four layers and the lower four layers.}
  \label{fig:equivalent-alternatives}
\end{figure*}

In Section~\ref{section:5.2}, we empirically demonstrate that the shared-parameter model generalizes well on both Test-IO and Test-OO. We further explain its effectiveness: sharing the parameters of the lower and upper four layers forces the model to align the embedding-level representations and hidden-state representations of atomic facts, while naturally transferring the reasoning capability learned in the lower layers to the upper layers. As a result, even when the model is only exposed to embedding-based reasoning patterns of out-of-distribution atomic facts during training, it can still perform reasoning over the hidden representations of out-of-distribution atomic facts at test time.

As a complement to Section~\ref{section:5.2}, in this section we explore alternative mechanisms that can serve as equivalent substitutes for the shared-parameter model, with the aim of gaining a deeper understanding of the fundamental sources of its effectiveness.
Specifically, we investigate two alternative training strategies that aim to replicate the key effects of parameter sharing without explicitly tying parameters across layers, as illustrated in Figure~\ref{fig:equivalent-alternatives}.
The left panel of the figure corresponds to a strategy that trains only the upper layers using embedded atomic facts, while the right panel illustrates a strategy that trains both the upper and lower layers separately under the same embedded-atomic-fact supervision.

\subsection{Training Upper Layers with Embedded Atomic Facts}
\label{appendix:equivalent-alternatives-upper}

\begin{figure*}[t]
    \centering
    \begin{subfigure}{\textwidth}
        \centering
        \includegraphics[width=\textwidth]{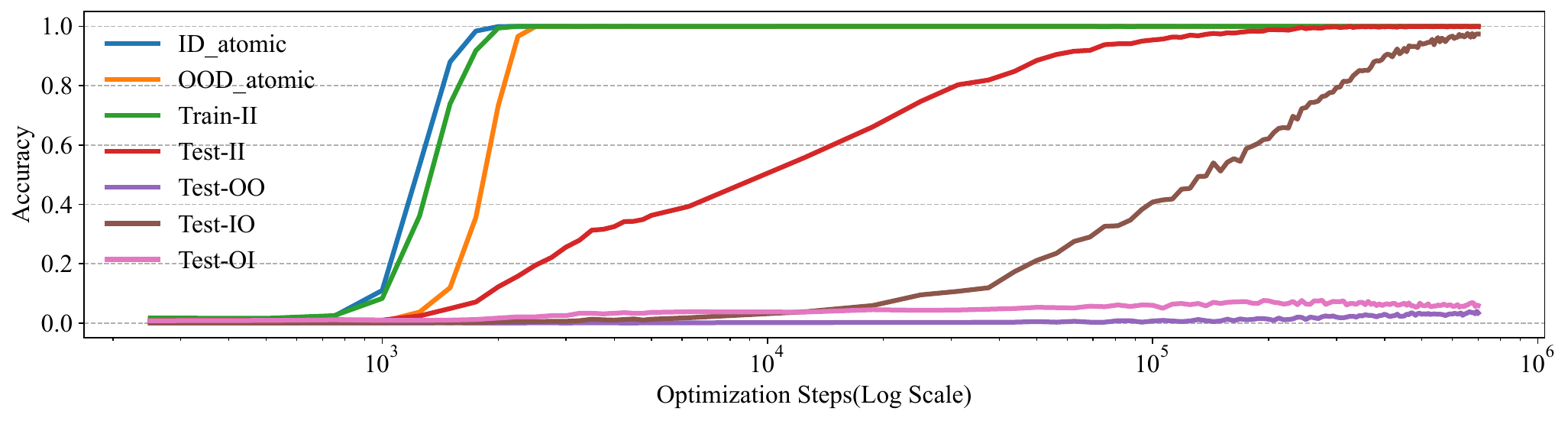}
        \caption{}  
        \label{fig:train_with_emb_patch_a}
    \end{subfigure}

    \vspace{0.5em}

    \begin{subfigure}{\textwidth}
        \centering
        \includegraphics[width=\textwidth]{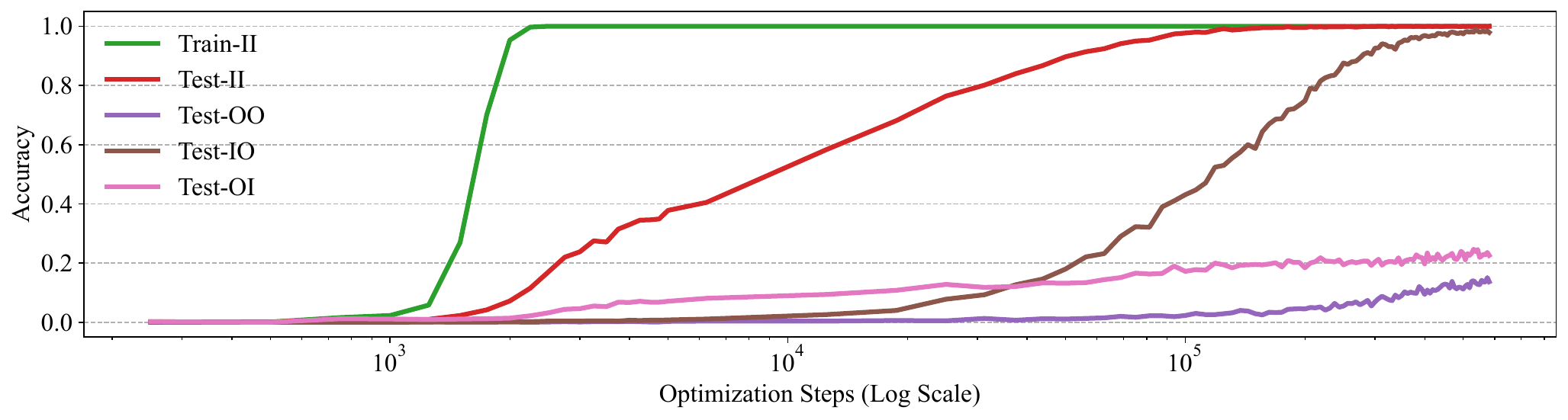}
        \caption{}  
        \label{fig:train_with_emb_patch_b}
    \end{subfigure}

    \caption{Training results for the two training strategies described in Appendix~\ref{appendix:equivalent-alternatives}:
    (a) the strategy corresponding to the left panel in Figure~\ref{fig:equivalent-alternatives}, where embeddings of atomic facts are additionally used to train the upper layers;
    (b) the strategy corresponding to the right panel in Figure~\ref{fig:equivalent-alternatives}, where embeddings of atomic facts are used to train both the upper and lower layers.
    Note that the accuracy on atomic facts is not shown in (b) because this training set does not include supervision that uses all layers to answer atomic facts.}
    \label{fig:train_with_emb_patch}
\end{figure*}

In this section, we augment the standard training pipeline with an additional procedure. After each training epoch, we further train the upper four layers of the model using all atomic facts, including both in-distribution (ID) and out-of-distribution (OOD) atomic facts. Specifically, we directly feed the embeddings of atomic facts $(E_e, E_r)$ into the fifth layer of the model, and supervise the model outputs using the target entities of these atomic facts. This procedure is designed to simulate a key property of the shared-parameter model, namely, enabling the upper layers to perform embedding-based reasoning.

The training results are shown in Figure~\ref{fig:train_with_emb_patch_a}. Compared with the standard training setting, the results differ in the following aspects:
\begin{itemize}
    \item \textbf{Test-IO and Test-OO.} The augmented training procedure enables strong generalization on Test-IO, achieving nearly 100\% accuracy, and weak but non-zero generalization on Test-OO, with around 4\% accuracy. In contrast, under the standard training setting, the model achieves almost zero accuracy on both Test-IO and Test-OO.
    
    \item \textbf{Test-OI.} Despite the improvements on Test-IO and Test-OO, the model only achieves limited generalization on Test-OI, with around 6\% accuracy. By comparison, the standard training procedure yields over 80\% accuracy on Test-OI.
\end{itemize}

We interpret these results as follows. Training the upper layers with embedding-based atomic facts reshapes their parameters and forces the intermediate representations of entities and relations to align with their corresponding embeddings. Moreover, since the upper layers are explicitly trained on OOD atomic facts, the model is able to generalize to Test-IO. However, introducing this additional training procedure also alters the parameters of the upper layers, which interferes with the model's ability to generalize on Test-OI.

Despite this limitation, the experiment remains informative. The fact that this training strategy enables generalization on Test-IO supports our hypothesis that the effectiveness of the shared-parameter model stems from endowing the upper layers with embedding-based reasoning capability.

\subsection{Training Upper and Lower Layers Separately with Embedded Atomic Facts}
\label{appendix:equivalent-alternatives-upper-lower}

In this section, we attempt to address the issue of disrupted generalization on Test-OI observed in Appendix~\ref{appendix:equivalent-alternatives-upper}. 
According to our analysis in Section~\ref{section:4.2}, the model’s ability to generalize on Test-OI arises from the consistency of representations for the same entity across different contexts. Furthermore, as analyzed by \citet{ye2026transformers}, the aggregation and alignment of representations generated from out-of-distribution atomic facts are enabled by the presence of in-distribution atomic facts that share the same target entity.
In the training strategy adopted in Appendix~\ref{appendix:equivalent-alternatives-upper}, the upper layers of the model are trained on atomic facts in two different phases: once through the standard dataset training and once through the additional embedding-based training. We hypothesize that this repeated training of atomic facts in different forms interferes with the model’s ability to generalize on Test-OI.

To test this hypothesis, we decouple the standard atomic-fact training from the additional embedding-based training. Specifically, in each training epoch, we train the lower four layers and the upper four layers separately using all atomic facts, while removing all atomic facts from the original training set and retaining only Train-II. The training results are shown in Figure~\ref{fig:train_with_emb_patch_b}. Compared with Appendix~\ref{appendix:equivalent-alternatives-upper}, we observe that, while maintaining high accuracy on Test-IO, the model achieves 25\% accuracy on Test-OI and 15\% accuracy on Test-OO. Both results are approximately four times higher than those reported in Appendix~\ref{appendix:equivalent-alternatives-upper}. These results support our hypothesis that the degradation in Test-OI generalization is indeed caused by interference introduced by the additional training procedure.

Although the proposed training strategy does not reach the accuracy achieved by the shared-parameter model on Test-OI and Test-OO (80\% and 60\%, respectively), it nonetheless attains a substantial level of performance and provides a closer functional approximation to shared-parameter training. Compared with the setting in Appendix~\ref{appendix:equivalent-alternatives-upper}, this experiment more strongly suggests that the success of shared-parameter training stems from enabling both the upper and lower layers of the model to perform embedding-based reasoning over atomic facts.

\section{Effect of Model Scale on the Training Performance of Looped Transformers}
\label{appendix:scale}

\begin{figure*}[t]
    \centering
    \includegraphics[width=\textwidth]{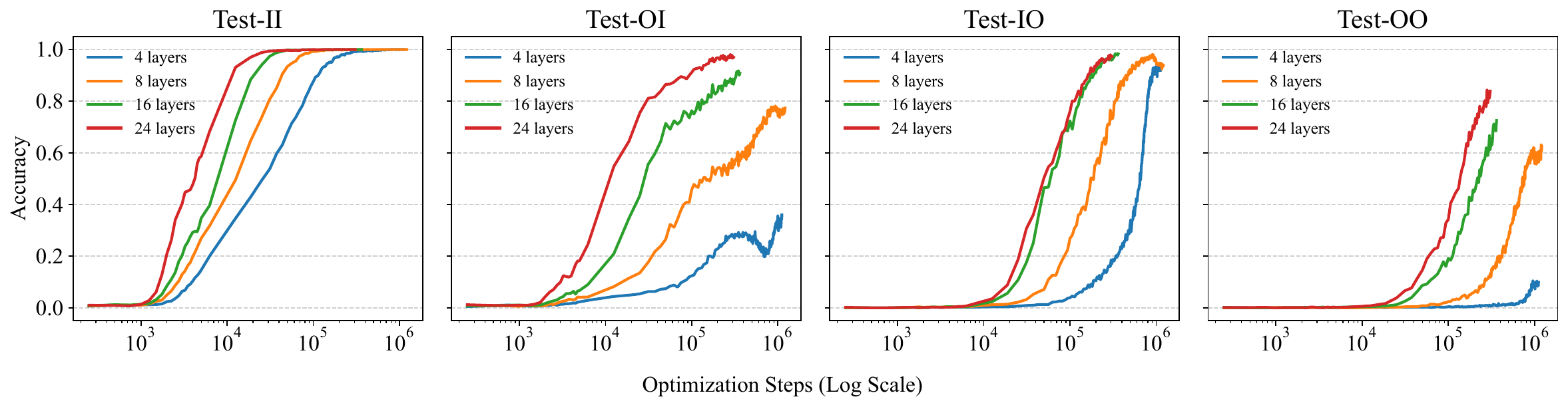}
    \caption{Test accuracy during training on looped transformers with varying model scale, controlled by the number of layers.}
    \label{fig:scale_accuracy}
\end{figure*}

In this section, we investigate how model scale affects the training dynamics of looped transformers by varying the number of layers. All models follow the GPT2 architecture with a hidden size of 768. We adopt the looped transformer design, where the parameters of the upper and lower halves of the network are shared across layers.

Specifically, we experiment with models of 4, 8, 16, and 24 layers. In the 4-layer setting, the parameters of the bottom 2 layers are shared with the top 2 layers. Similarly, the 8-layer model shares parameters between the bottom 4 and top 4 layers, and so on. The training results are shown in Figure~\ref{fig:scale_accuracy}.

\begin{figure*}[t]
    \centering
    \includegraphics[width=\textwidth]{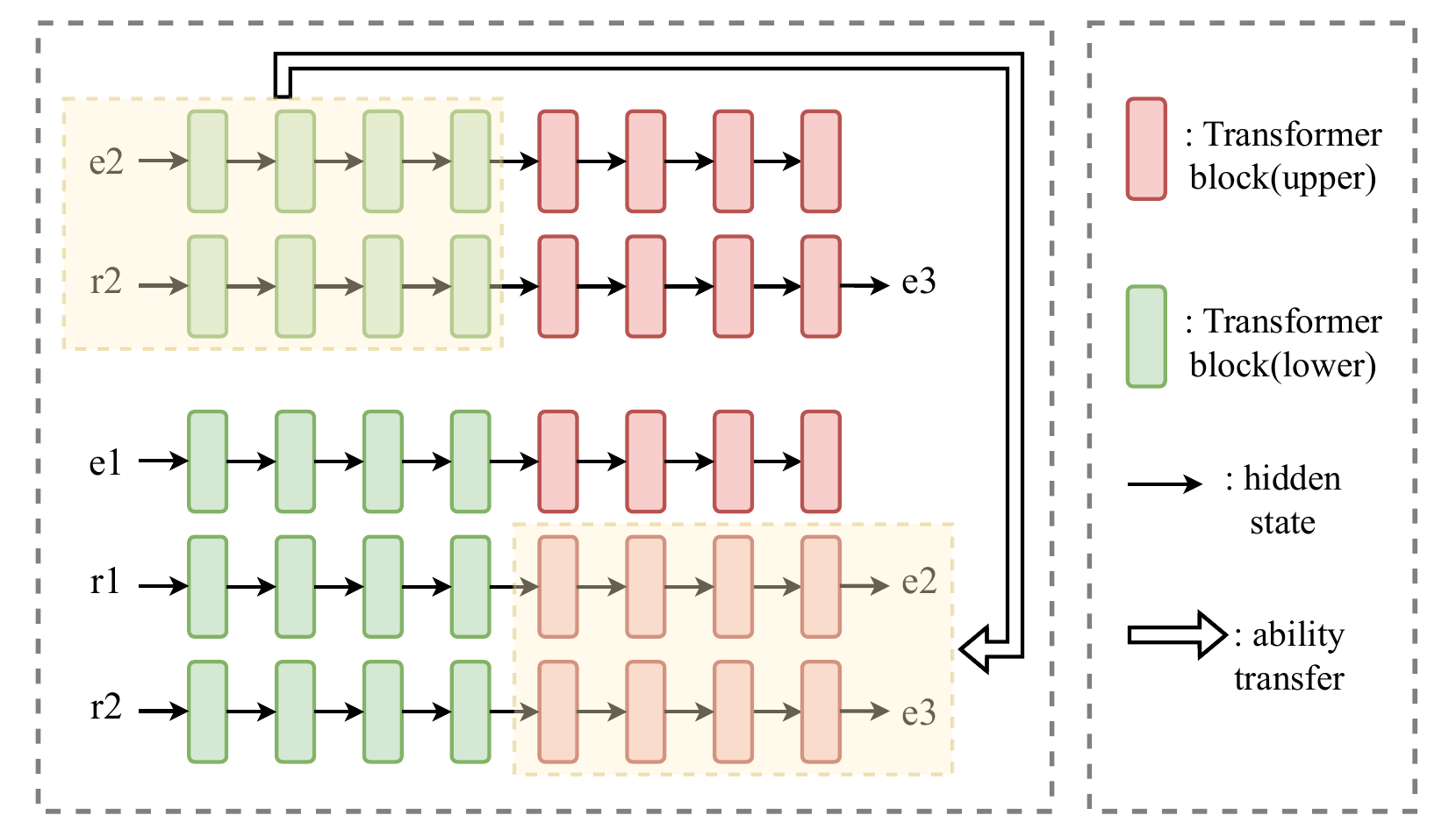}
    \caption{Visualization of how a model trained with a looped architecture transfers embedding-based atomic fact reasoning learned in lower layers to higher layers.
    The bottom part illustrates the reasoning process for a two-hop instance, while the top part shows the embedding-based reasoning process corresponding to the second hop of the same instance.}
    \label{fig:alignment}
\end{figure*}

From Figure~\ref{fig:scale_accuracy}, we observe the following trends:

\begin{itemize}
    \item \textbf{Generalization accuracy.}
    Looped transformers with different layer counts can all achieve nearly 100\% accuracy on Test-II and Test-IO. However, on Test-IO and Test-OO, models with more layers demonstrate better generalization performance. Models with fewer layers exhibit weaker generalization ability. In particular, the 4-layer model shows unstable training behavior and achieves only around 10\% accuracy on Test-OO. This phenomenon is likely due to the limited capacity of small-scale models, which is insufficient to support robust generalization.

    \item \textbf{Generalization speed.}
    Models with more layers tend to generalize faster during training. In addition, deeper models show more stable generalization dynamics, with smaller fluctuations in accuracy throughout the training process.
\end{itemize}

\section{Detailed Evidence on Alignment under Looped Training}
\label{appendix:Detailed Evidence on Alignment}

This section supplements Section~\ref{section:5.2}. We provide a detailed description of how the results in Section~\ref{section:5.2} are computed and present additional evidence for \emph{Representation--Input Alignment under Looped Training}.
In Section~\ref{section:5.2}, we only report the alignment results for
(i) $(h_4(r_1), h_4(r_2))$ vs.\ $(E_{e_2}, E_{r_2})$, and
(ii) $(h_5(r_1), h_5(r_2))$ vs.\ $(h_1(e_2), h_1(r_2))$.
Here, we extend this analysis by reporting cosine similarities between hidden states across a broader range of layer pairs.

We first describe how the results in Section~\ref{section:5.2} are computed.
For each two-hop instance from the training set (Train-II), we identify the atomic fact corresponding to its second hop, which is in-distribution (ID).
For each two-hop instance from the test set (Test-IO), we identify the atomic fact corresponding to its second hop, which is out-of-distribution (OOD).
This matching strategy allows us to examine whether the model reuses embedding-level representations of atomic facts when performing the second hop of reasoning, as illustrated in Figure~\ref{fig:alignment}.

Specifically, for a two-hop query $\langle e_1, r_1, r_2 \rangle$, we extract the hidden states at the position of $r_1$ from all Transformer layers, and compare them with the hidden states at the position of $e_2$ from the corresponding second-hop atomic fact $\langle e_2, r_2 \rangle$.
We compute a layer-wise cosine similarity matrix of size $(L+1)\times(L+1)$, where $L$ denotes the number of Transformer layers\footnote{Layer~0 corresponds to the input embedding without positional encoding.}.
Similarly, we extract the hidden states at the position of $r_2$ in $\langle e_1, r_1, r_2 \rangle$ and compare them with the hidden states at the position of $r_2$ in the atomic fact $\langle e_2, r_2 \rangle$, again computing a $(L+1)\times(L+1)$ cosine similarity matrix.

Finally, we average the cosine similarity matrices over all matched two-hop instances to obtain the final mean cosine similarity matrices.
We present the averaged similarity matrices separately for Train-II and Test-IO, and compare models trained with the looped architecture against normally trained models.
Figure~\ref{fig:layerwise_similarity} visualizes the resulting mean cosine similarity matrices.

From Figure~\ref{fig:layerwise_similarity}, it is clear that for models trained under the looped architecture (left column), the cosine similarity matrices at positions $r_1$ and $r_2$ for both Train-II and Test-IO ((a), (c), (e), and (g)) show significantly higher values along the diagonal directions within the red-square-highlighted regions.
Notably, the entries at position (1,5) (row, column) exhibit particularly strong similarity across all subfigures. In contrast, models trained in the standard manner do not display this pattern. This indicates that for models trained under the looped architecture, the implicit second-hop reasoning in the two-hop queries is aligned with the reasoning over the corresponding single-hop atomic facts (based on embeddings), achieving strong alignment at the fifth layer. Under this input alignment condition, the model can naturally transfer the reasoning ability over out-of-distribution atomic facts learned in lower layers to higher layers, enabling effective generalization to both Test-IO and Test-OO.

\begin{figure*}[t]
    \centering

    \begin{subfigure}{0.45\textwidth}
        \includegraphics[width=\linewidth]{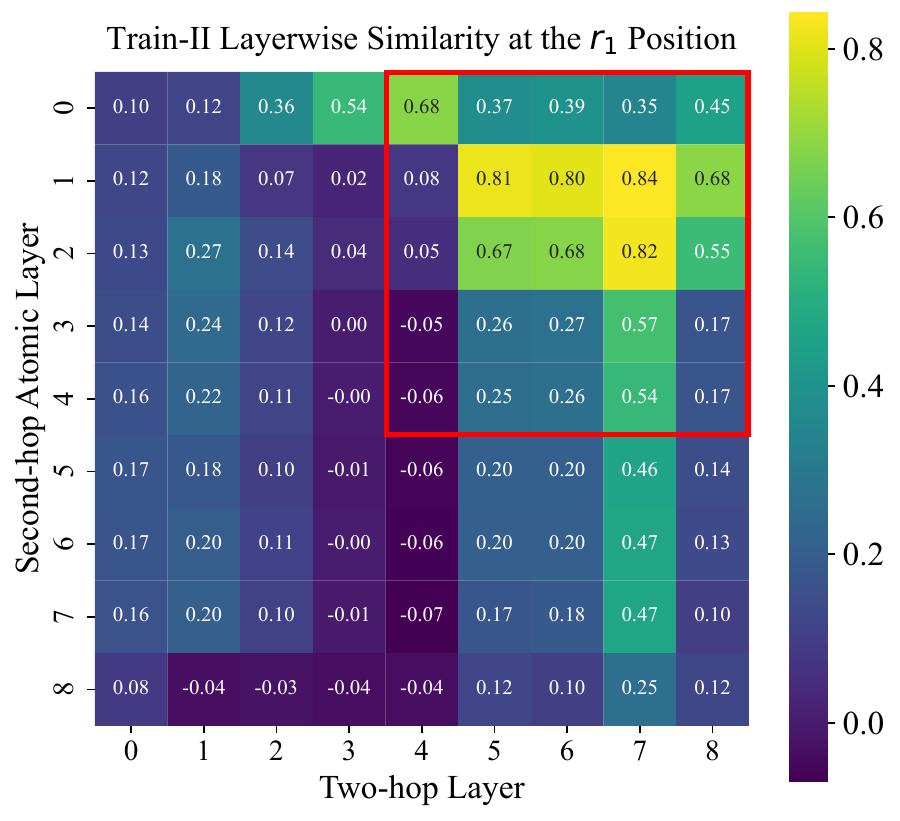}
        \caption{}
    \end{subfigure}
    \hfill
    \begin{subfigure}{0.45\textwidth}
        \includegraphics[width=\linewidth]{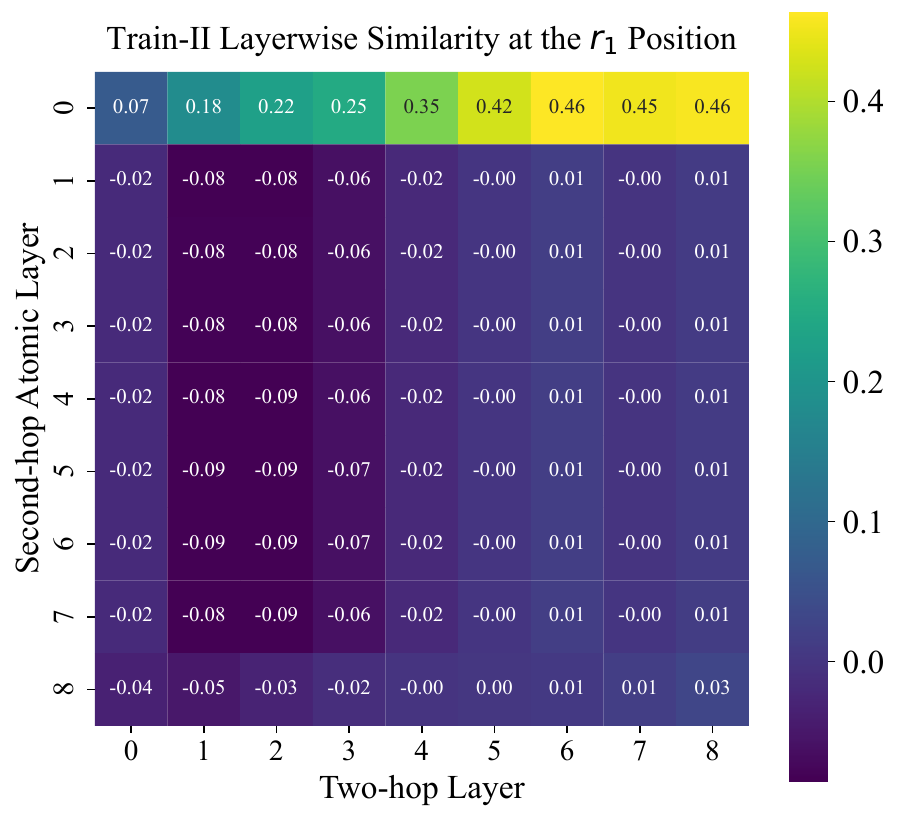}
        \caption{}
    \end{subfigure}

    \vspace{0.6em}

    \begin{subfigure}{0.45\textwidth}
        \includegraphics[width=\linewidth]{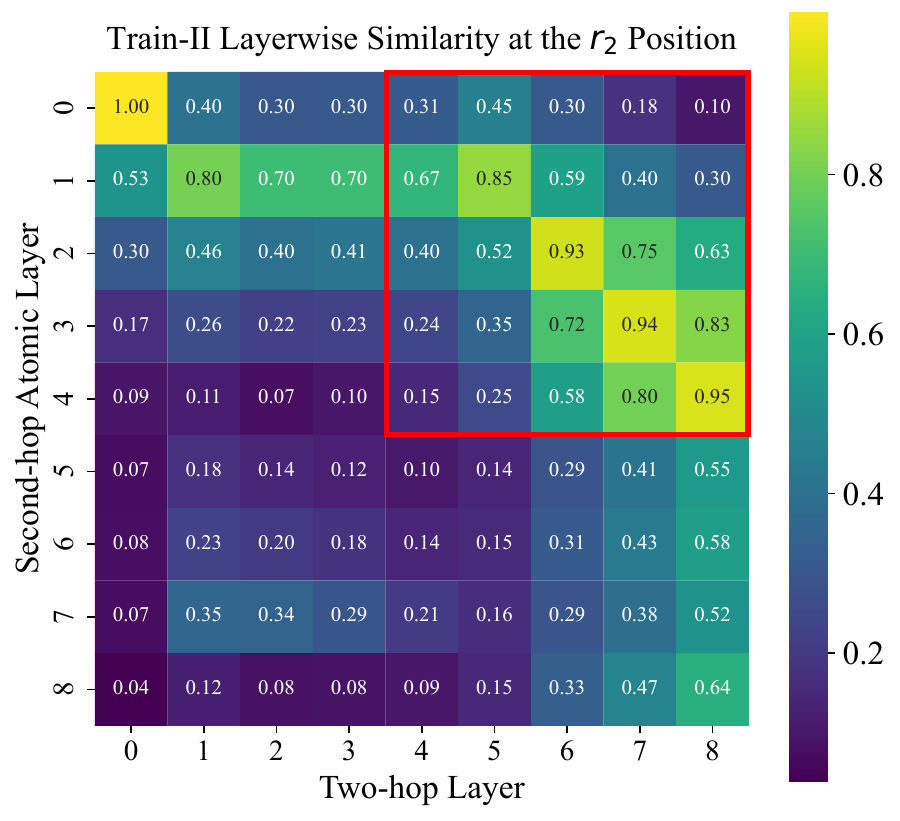}
        \caption{}
    \end{subfigure}
    \hfill
    \begin{subfigure}{0.45\textwidth}
        \includegraphics[width=\linewidth]{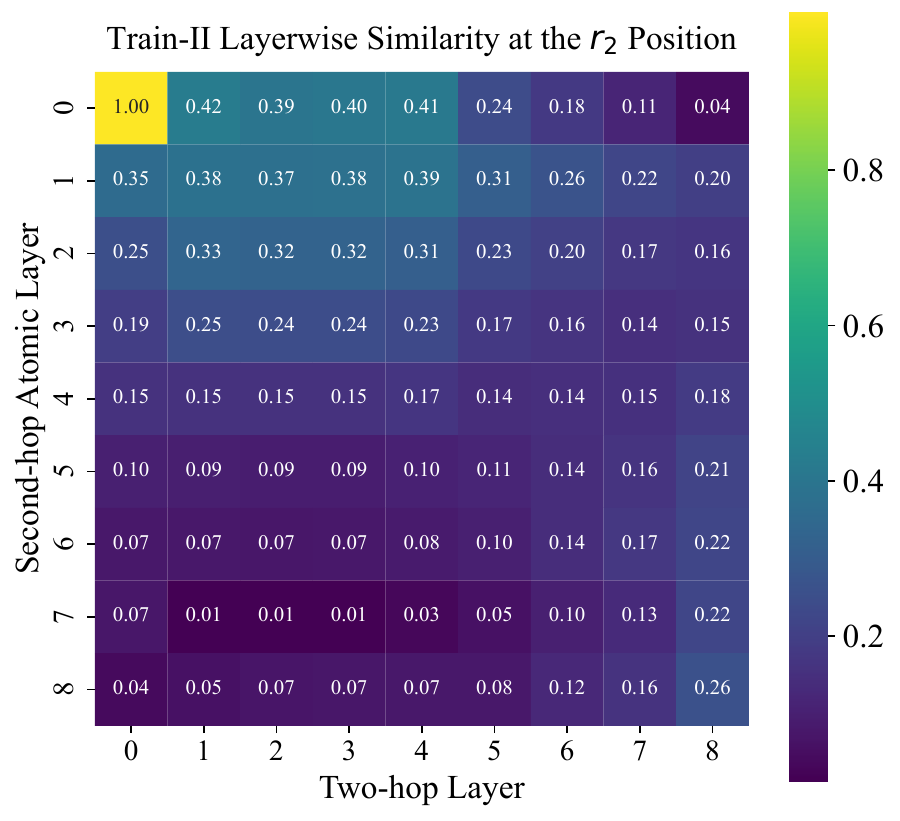}
        \caption{}
    \end{subfigure}

    \vspace{0.6em}

    \begin{subfigure}{0.45\textwidth}
        \includegraphics[width=\linewidth]{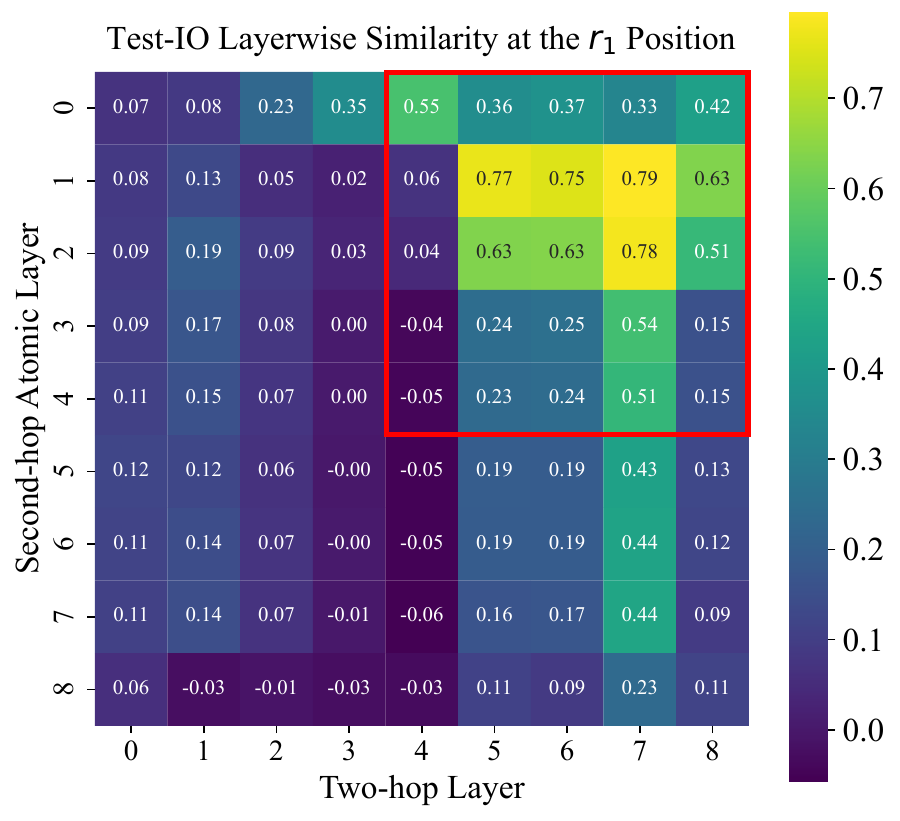}
        \caption{}
    \end{subfigure}
    \hfill
    \begin{subfigure}{0.45\textwidth}
        \includegraphics[width=\linewidth]{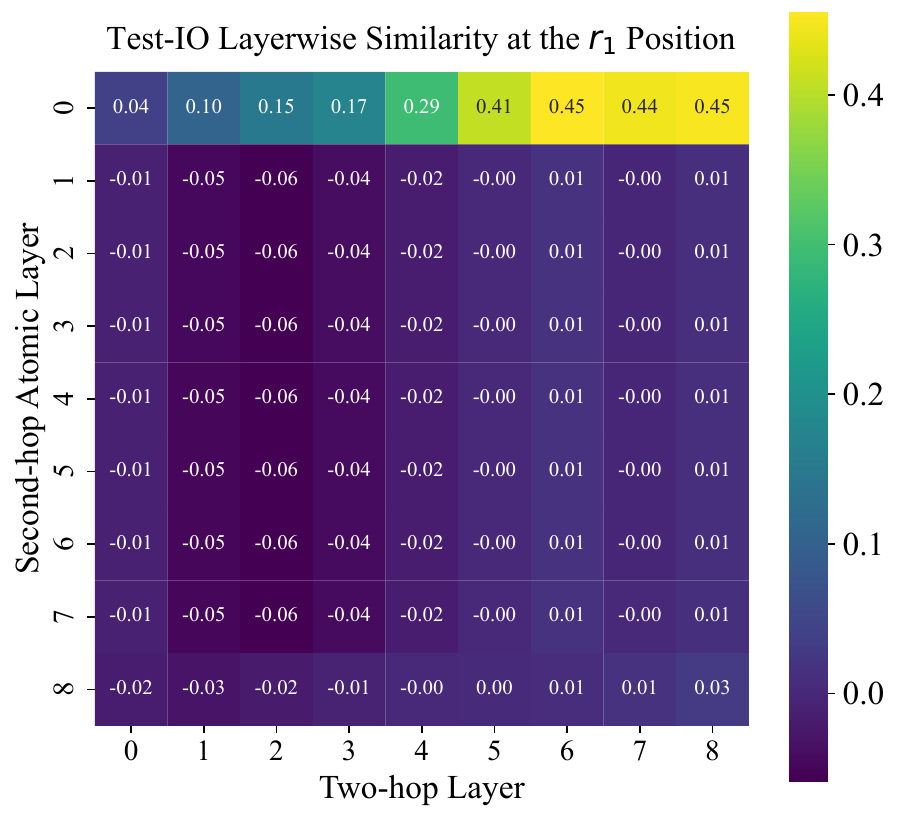}
        \caption{}
    \end{subfigure}

    \caption*{}
\end{figure*}
\begin{figure*}[t]
    \ContinuedFloat
    \centering

    \begin{subfigure}{0.45\textwidth}
        \includegraphics[width=\linewidth]{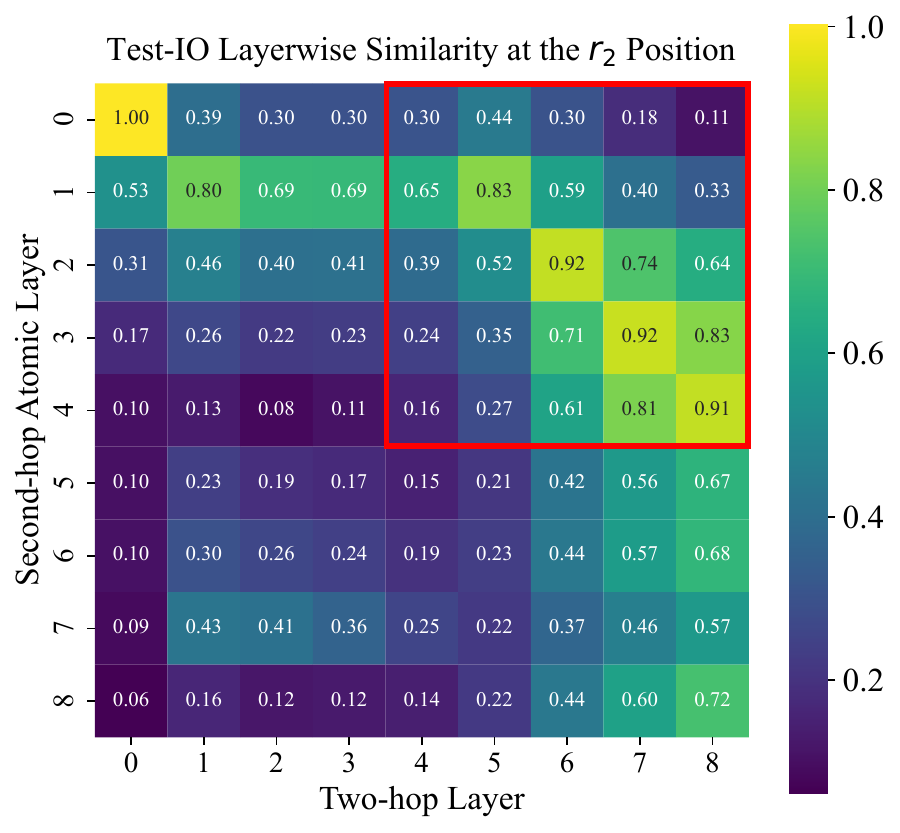}
        \caption{}
    \end{subfigure}
    \hfill
    \begin{subfigure}{0.45\textwidth}
        \includegraphics[width=\linewidth]{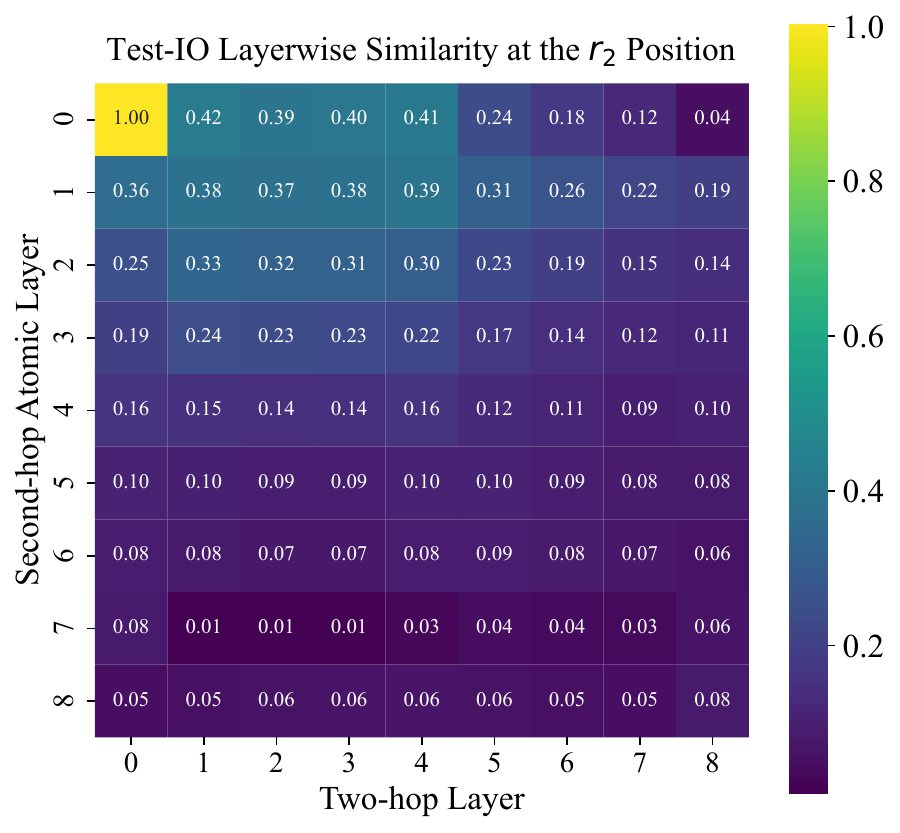}
        \caption{}
    \end{subfigure}

    \caption{
    Layerwise cosine similarity heatmaps comparing models trained under a recurrent architecture (left column) and models trained in the standard manner (right column).
    For each two-hop query $\langle e_1, r_1, r_2 \rangle$, hidden states at the position of $r_1$ are compared with the corresponding hidden states at $e_2$ in the atomic fact $\langle e_2, r_2 \rangle$, and hidden states at the position of $r_2$ are compared with the hidden states at $r_2$ in the atomic fact.
    Subfigures (a) and (b) show Train-II at position $r_1$,
    (c) and (d) show Train-II at position $r_2$,
    (e) and (f) show Test-IO at position $r_1$,
    and (g) and (h) show Test-IO at position $r_2$.
    }
    \label{fig:layerwise_similarity}
\end{figure*}

\section{Extraction of a Real-World Two-Hop Dataset}
\label{appendix:real_two_hop}

To evaluate the effectiveness of looped transformers on real-world two-hop reasoning, we construct a dataset from Wikidata5M by extracting a dense subgraph.

Prior work has shown that successful generalization on two-hop reasoning critically depends on the ratio between compositional training samples (Train-II) and atomic facts. \citet{wang2024grokking} identify a threshold phenomenon: models exhibit grokking behavior on Test-II only when this ratio, denoted as $\phi$, exceeds approximately 5.4. However, real-world multi-hop datasets are typically too sparse to satisfy this condition. For example, \citet{grokking_in_the_wild} report that the graph induced by 2WikiMultiHopQA has $\phi \approx 0.5$, which is insufficient for grokking to emerge. To address this issue, \citet{grokking_in_the_wild} increase $\phi$ by augmenting the knowledge graph with additional edges, reaching $\phi = 6.25$ and enabling generalization on Test-II, although the model still fails to generalize to Test-OO. In our work, instead of augmenting the dataset, we directly extract a dense subgraph from Wikidata5M, which provides a simple and controlled way to increase $\phi$ while focusing on evaluating the effectiveness of looped transformers on real-world data.

\subsection{Subgraph Construction}

We construct the subgraph from Wikidata5M by focusing on entities of type \textit{human}. We first collect all such entities and compute their in-degrees and out-degrees within the original graph. Each entity is then assigned a score:
\[
\text{score}(e) = \frac{\text{in-degree}(e) \times \text{out-degree}(e)}{\text{in-degree}(e) + \text{out-degree}(e)}.
\]
This scoring function favors entities that simultaneously have high in-degree and out-degree, i.e., entities that are well-connected both as sources and as targets. Intuitively, such entities are more likely to serve as intermediate nodes in two-hop paths, thereby contributing to a denser set of compositional relations.

We rank entities by this score and select the top 100 as central nodes. The subgraph is then constructed by including these entities and all their neighboring entities as nodes, along with all edges among them. This yields a set of single-hop facts and all candidate two-hop paths within the extracted subgraph.

\subsection{Two-Hop Path Filtering and Balancing}

We further refine the two-hop samples to mitigate shortcut learning and encourage genuine compositional reasoning.

First, we remove cyclic paths where the head and tail entities are identical.
Second, we control the target distribution for each relation pair $(r_1, r_2)$. Without such control, certain targets may dominate a given relation pair, allowing the model to exploit spurious correlations between $(r_1, r_2)$ and $e_3$ rather than performing two-hop reasoning.
Concretely, for each $(r_1, r_2)$, we group paths by their target entity $e_3$ and enforce a maximum proportion constraint. Let $N$ denote the total number of retained paths for this pair. We ensure that no single target entity accounts for more than $0.2N$. In addition, we discard relation pairs with fewer than 10 distinct target entities to ensure sufficient diversity.

 After filtering, the ratio of Train-II paths to single-hop samples ($\phi$) is 9.60. For the actual training set, we sample Train-II paths to achieve $\phi = 7.2$, providing a dense yet controlled setting for evaluating the effectiveness of looped transformers on real-world two-hop reasoning. The statistics of the resulting dataset are summarized in Table~\ref{tab:real_data_stats}.

\begin{table}[H]
\centering
\small
\begin{tabular}{c c c}
\hline
\textbf{Category} & \textbf{Data Type} & \textbf{Quantity} \\
\hline
\multirow{2}{*}{Graph Statistics} 
    & Entities  & 14,995 \\
    & Relations & 176 \\
\hline
\multirow{3}{*}{Training Set} 
    & ID\_atomic  & 15,375 \\
    & OOD\_atomic & 809 \\
    & Train-II    & 110,700 \\
\hline
\multirow{7}{*}{Test Set} 
    & ID\_atomic  & 1,000 \\
    & OOD\_atomic & 1,000 \\
    & Train-II    & 1,000 \\
    & Test-II     & 1,000 \\
    & Test-IO     & 1,000 \\
    & Test-OI     & 1,000 \\
    & Test-OO     & 480 \\
\hline
\end{tabular}
\caption{Statistics of the extracted two-hop dataset and subgraph. The graph statistics include the total number of entities and relations. Training and test splits are further broken down by sample type.}
\label{tab:real_data_stats}
\end{table}

\section{Comparison with Explicit CoT Training}
\label{appendix:cot_training}

Although our work focuses on implicit reasoning, it is necessary to compare our approach with explicit Chain-of-Thought (CoT) supervision. Prior work by \citet{yaoUnveilingMechanismsExplicit2025}, using the same dataset setting introduced by \citet{wang2024grokking}, demonstrated that explicit CoT supervision can substantially improve out-of-distribution two-hop generalization, achieving nearly 100\% accuracy on the Test-OO split.

In contrast, our proposed solution, the looped transformer, does not achieve perfect Test-OO accuracy. However, it addresses a fundamental architectural limitation of implicit reasoning models: the upper layers of standard transformers lack robust out-of-distribution reasoning capabilities. Moreover, explicit CoT supervision requires constructing additional supervision signals for intermediate entities, which introduces extra annotation procedures and training overhead. Our approach, by comparison, only requires modifying the model architecture without introducing additional supervision.

\end{document}